\documentclass[preprint]{vgtc}               % preprint

\preprinttext{To appear in the Proceedings of the 19th IEEE Pacific Visualization Conference (PacificVis 2026), Sydney, Australia.}

\ifpdf%                                % if we use pdflatex
  \pdfoutput=1\relax                   % create PDFs from pdfLaTeX
  \usepackage{graphicx}                % allow us to embed graphics files
  \DeclareGraphicsExtensions{.pdf,.png,.jpg,.jpeg} % for pdflatex we expect .pdf, .png, or .jpg files
\else%                                 % else we use pure latex
  \usepackage{graphicx}                % allow us to embed graphics files
  \DeclareGraphicsExtensions{.eps}     % for pure latex we expect eps files
\fi%

\graphicspath{{figures/}{pictures/}{images/}{./}} % where to search for the images

\usepackage{microtype}                 % use micro-typography (slightly more compact, better to read)
\PassOptionsToPackage{warn}{textcomp}  % to address font issues with \textrightarrow
\usepackage{textcomp}                  % use better special symbols
\usepackage{mathptmx}                  % use matching math font
\usepackage{times}                     % we use Times as the main font
\usepackage{cite}                      % needed to automatically sort the references
\usepackage{tabu}                      % only used for the table example
\usepackage{xcolor} 
\usepackage{booktabs}                  % only used for the table example

\newcommand\blfootnote[1]{%
  \begingroup
  \renewcommand\thefootnote{}\footnote{#1}%
  \addtocounter{footnote}{-1}%
  \endgroup
}

\onlineid{6564}

\vgtccategory{Research}

\vgtcinsertpkg

\title{REV-INR: Regularized Evidential Implicit Neural Representation for Uncertainty-Aware Volume Visualization}

\author{Shanu Saklani\\ %
        \scriptsize INSIGHT Lab., IIT Kanpur, India %
\and Tushar M. Athawale\\ %
     \scriptsize Oak Ridge National Laboratory, USA %
\and Nairita Pal\\ %
     \scriptsize IIT Kharagpur, India %
\and David Pugmire\\ %
     \scriptsize Oak Ridge National Laboratory, USA %
\and Christopher R. Johnson\\ %
     \scriptsize University of Utah, USA
\and Soumya Dutta\thanks{e-mail: soumyad@cse.iitk.ac.in (Corresponding author.)}\\ %
     \scriptsize INSIGHT Lab., IIT Kanpur, India} %

\abstract{Applications of Implicit Neural Representations (INRs) have emerged as a promising deep learning approach for compactly representing large volumetric datasets. These models can act as surrogates for volume data, enabling efficient storage and on-demand reconstruction via model predictions. However, conventional deterministic INRs only provide value predictions without insights into the model's prediction uncertainty or the impact of inherent noisiness in the data. This limitation can lead to unreliable data interpretation and visualization due to prediction inaccuracies in the reconstructed volume. Identifying erroneous results extracted from model-predicted data may be infeasible, as raw data may be unavailable due to its large size. To address this challenge, we introduce \textbf{\textit{REV-INR}}, \boldunderline{R}egularized \boldunderline{Ev}idential \boldunderline{I}mplicit \boldunderline{N}eural \boldunderline{R}epresentation, which learns to predict data values accurately along with the associated coordinate-level data uncertainty and model uncertainty using only a single forward pass of the trained REV-INR during inference. By comprehensively comparing and contrasting REV-INR with existing well-established deep uncertainty estimation methods, we show that REV-INR achieves the best volume reconstruction quality with robust data (aleatoric) and model (epistemic) uncertainty estimates using the fastest inference time. Consequently, we demonstrate that REV-INR facilitates assessment of the reliability and trustworthiness of the extracted isosurfaces and volume visualization results, enabling analyses to be solely driven by model-predicted data.
} % end of abstract

\graphicspath{{figs/}{figures/}{pictures/}{images/}{./}} % where to search for the images

\CCScatlist{
  \CCScatTwelve{Computing methodologies}{Machine learning}{}{};
  \CCScatTwelve{Human-centered computing}{Visualization}{}{};
  \CCScatTwelve{Mathematics of computing}{Probability and statistics}{}{};
  \CCScatTwelve{Computing methodologies}{Computer graphics}{}{}
}

\usepackage{tabu}                      % only used for the table example
\usepackage{booktabs}                  % only used for the table example
\usepackage{lipsum}                    % used to generate placeholder text
\usepackage{mwe}                       % used to generate placeholder figures
\usepackage{color}
\usepackage{subcaption}
\usepackage{amsmath}
\usepackage{amsfonts}
\usepackage{graphicx}  
\usepackage{amssymb}
\usepackage{multirow}
\usepackage[ruled,vlined]{algorithm2e}
\usepackage{mathrsfs}
\usepackage[normalem]{ulem} %% For underlining and bold faced text
\newcommand{\boldunderline}[1]{\textbf{\uline{#1}}}

\usepackage{makecell}
\DeclareUnicodeCharacter{2212}{\ensuremath{-}}

\useunder{\uline}{\ul}{}

\newcommand{\rmark}[1]{#1}

\newenvironment{tight_enumerate}{
\begin{enumerate}
  \setlength{\itemsep}{0pt}
  \setlength{\parskip}{4.5pt}
}{\end{enumerate}}

\usepackage{mathptmx}                  % use matching math font

\begin{document}

%%%%%%%%%%%%%%%%%%%%%%%%%%%%%%%%%%%%%%%%%%%%%%%%%%%%%%%%%%%%%%%%
%%%%%%%%%%%%%%%%%%%%%% START OF THE PAPER %%%%%%%%%%%%%%%%%%%%%%
%%%%%%%%%%%%%%%%%%%%%%%%%%%%%%%%%%%%%%%%%%%%%%%%%%%%%%%%%%%%%%%%

\firstsection{Introduction}
\maketitle

Implicit neural representations (INRs) have gained wide popularity in the visualization community as compact surrogates for large volumetric datasets~\cite{levine_neural_compression, coordnet, han2020ssr, stsrinr, kdinr}. INRs effectively learn complex scalar fields and enable on-demand volume prediction, allowing downstream tasks such as isosurface extraction and volume rendering to be performed directly on INR-predicted data. Although INRs demonstrate high-quality reconstruction, prediction inaccuracies are inevitable. In real-world scenarios, where raw high-resolution data may not be stored due to its size~\cite{Dutta_TVCG_DL_Uncet, vectorNet, Tianyu}, error estimation becomes infeasible, making it difficult to assess the quality, reliability, and trustworthiness of the results. Most INR-based works in scientific visualization~\cite{levine_neural_compression, coordnet, han2020ssr, stsrinr, kdinr} employ deterministic models that predict only values, without quantifying uncertainty. To address this  limitation, recent researchers have started exploring uncertainty-aware INRs that predict both data values and associated model-level uncertainties~\cite{uncertVolren, Dutta_TVCG_DL_Uncet, Tianyu, Jingyi_SurroFlow}. So far, these efforts primarily examine model (epistemic) uncertainty, while the quantification, application, and usability of data-level (aleatoric) uncertainty has not been explored yet. For generating high-quality, reliable, and trustworthy visualizations, comprehensive treatment of both epistemic and aleatoric uncertainties is essential~\cite{Kendall_UQ}. Moreover, since uncertainty is an inherent model property that can be estimated without raw data, it provides a powerful mechanism for producing informative and trustworthy results in scientific applications.

A widely adopted approach to estimate uncertainty in INRs is to use an ensemble of INRs and treat their prediction variance as the uncertainty estimate. While such deep ensembles~\cite{deepEnsembles} are often found to be very powerful~\cite{DENS_superior1, DENS_superior2, DENS_superior3, uncertVolren}, their strength comes at the cost of excessively high training times and increased storage requirements, since multiple models must be trained and stored~\cite{uncertVolren, Dutta_TVCG_DL_Uncet}. Therefore, in this work, we focus on developing \textit{single-model-based uncertainty-aware INRs}. To the best of our knowledge, this is the first work that quantify and study both model-level (epistemic) and data-level (aleatoric) uncertainty estimates using a single INR, and demonstrates their interpretation, usability, and applicability in INR-driven volume visualization tasks. Model uncertainty helps domain experts identify regions where the INR is over- or under-confident in its predictions, while data uncertainty highlights the irreducible noise (stochasticity) in the data and potential regions where such noise may affect analysis results and visualizations. Thus, a clear separation, estimation, and visual exploration of these two fundamental uncertainty types are essential for effectively evaluating INR reliability and understanding the influence of inherent data noise~\cite{Kendall_UQ}.\blfootnote{This manuscript has been authored by UT-Battelle, LLC under Contract No. DE-AC05-00OR22725 with the U.S. Department of Energy. The publisher, by accepting the article for publication, acknowledges that the U.S. Government retains a non-exclusive, paid up, irrevocable, world-wide license to publish or reproduce the published form of the manuscript, or allow others to do so, for U.S. Government purposes. The DOE will provide public access to these results in accordance with the DOE Public Access Plan (\url{http://energy.gov/downloads/doe-public-access-plan}).}

To comprehensively quantify and evaluate the significance of epistemic and aleatoric uncertainties in volume visualization, we propose REV-INR, a regularized evidential implicit neural representation (INR) that produces high-quality data predictions along with reliable epistemic and aleatoric uncertainties — all within a single forward pass at inference time. By employing evidential learning~\cite{evidential_survey_arxiv}, REV-INR learns the parameters of a higher-order posterior distribution, thereby providing closed-form solutions for predicting data values together with both types of uncertainty. To produce interpretable uncertainty estimates, we propose two novel uncertainty regularization techniques for REV-INR. To compare REV-INR with existing single-model-based approaches, we develop two additional uncertainty-aware INRs. The first, referred to as \textit{MCD-INR}, is based on the Monte Carlo Dropout approach~\cite{gagh16, uncertVolren, Tianyu}. Since conventional MCD-INR only estimates epistemic uncertainty, we extend MCD-INR to also capture data (aleatoric) uncertainty. The second model is adapted from  RMDSRN~\cite{Tianyu} which is  also designed to predict only epistemic uncertainty, we build a new INR, consisting of a shared encoder and multiple decoders, similar to  RMDSRN, to predict both epistemic and aleatoric uncertainties and refer to this new regularized model as \textit{RMD-INR}. 

To enable reliable and robust analysis of volume data using INR predictions, we comprehensively evaluate REV-INR against MCD-INR and RMD-INR. We first assess reconstruction quality and the accuracy of volume visualization results. Next, we employ uncertainty-aware isosurface visualization~\cite{Athawale_UncertaintyMarchingCubes, Pothkow2011} to interpret and assess the uncertainty estimates. By analyzing isosurface structures and their topology, experts can better understand scalar field topology~\cite{topology_need4, topology_need2}. Thus, it is essential that INR-generated data-driven isosurfaces preserve true structures and enable the identification of regions where the INR is under- or over-confident and prone to errors. We further evaluate the interpretability and usability of both uncertainty types by examining their correlations with prediction error, local data variance, and interpolation-based errors. Additionally, we provide results from a deterministic INR for completeness. Our comprehensive evaluations show that REV-INR outperforms MCD-INR and RMD-INR, achieving superior reconstruction quality, visualization results, reliable uncertainty estimates, and the fastest inference times. Hence, our contributions are as follows:   
\begin{tight_enumerate}
\item We propose REV-INR, a new uncertainty-aware Regularized Evidential INR with novel uncertainty regularization methods that enables efficient learning of large volumetric data while delivering fast, high-quality inference and fine-grained prediction quality assessment through coordinate-level epistemic and aleatoric uncertainty estimates. 
\item We (a) extend Monte Carlo Dropout-INR, and (b) build another uncertain SRN, RMD-INR, adapted from RMDSRN, enabling the estimation of aleatoric (data) uncertainty in both models in addition to epistemic uncertainty.
\item We extensively compare and contrast REV-INR with MCD-INR, RMD-INR, and state-of-the-art compression methods to demonstrate its superiority in fast volume reconstruction, reliable and meaningful epistemic and aleatoric uncertainty estimation, and trustworthy, accurate visualization generation.
\end{tight_enumerate}

\section{Related Works}

\subsection{Techniques of Uncertain Volume Visualization}
Visualization of volumetric data with uncertainty is an important visualization task. A typical representation of such volumetric data is in the form of per-point probability distribution. Liu et al. use flickering to represent uncertainty in such volume data~\cite{Liu2012}. A statistical volume visualization framework is proposed in~\cite{stat_volren}. Athawale et al. investigate uncertainty visualization in volume rendering using nonparametric models~\cite{AthawaleAJE}. Pöthkow et al. develop a method to compute the level-crossing probability, which is further refined to determine the probability for each cell~\cite{51187755, Pothkow2011}. Whitaker et al.~\cite{6634129} examine uncertainty visualization in ensembles of contours. Athawale et al. extensively explore uncertainty in isosurface extraction~\cite{Athawale_1, Athawale_2}.

\subsection{Deep Learning for Scalar Field Visualization}
Deep learning methods for creating compact representations of scalar field data have been extensively explored in recent years~\cite{levine_neural_compression, kdinr, weiss2022fast}. The use of volume-rendered images for scalar data analysis has been demonstrated by Hong et al.~\cite{hong2019dnn}, He et al.~\cite{he2019insitunet}, and Berger et al.~\cite{berger2018generative}. For domain-knowledge-aware volume data compression, latent space techniques have been introduced in~\cite{shen2022idlat}. Advances in generating high-resolution spatiotemporal volumes have been reported in~\cite{han2020ssr, wurster2022deep, han2021stnet}. Deep neural networks (DNNs) have also been employed as surrogates for exploring parameter spaces in ensemble data~\cite{he2019insitunet, shi2022vdl}. Weiss et al.~\cite{weiss2019volumetric} perform isosurface visualization using DNNs, while Han et al.~\cite{han_dl_prob_MC} utilize deep learning to enhance the probabilistic marching cubes algorithm for isosurface extraction. 

Recently, researchers have begun investigating epistemic (model) uncertainty in visual analysis. Model uncertainty in CNN-based view synthesis has been studied in~\cite{Dutta_TVCG_DL_Uncet}, while epistemic uncertainty in super-resolution~\cite{PSRFlow} and parameter space exploration~\cite{Jingyi_SurroFlow} has also been explored. The impacts of epistemic uncertainty on INR-based volume visualization are analyzed in~\cite{uncertVolren}. Uncertainty-aware vector data modeling has been addressed in~\cite{vectorNet}. Uncertain-INR has been applied to represent uncertainty in CT data~\cite{uncertainINR} and remote sensing images~\cite{uncertINR_remotesense}. Furthermore, an uncertainty-aware regularized multi-decoder scene representation network (RMDSRN) for scalar data visualization has been proposed in~\cite{Tianyu}.

Our survey reveals that existing research has only focused on epistemic (model) uncertainty. However, a detailed investigation of data (aleatoric) uncertainty, as well as a comprehensive framework for jointly estimating both uncertainties in the context of volume modeling, remains unexplored — a gap that we aim to address.

\begin{figure*}[!tb]
\centering
\begin{subfigure}[t]{0.32\textwidth}
    \centering
    \includegraphics[width=\textwidth]{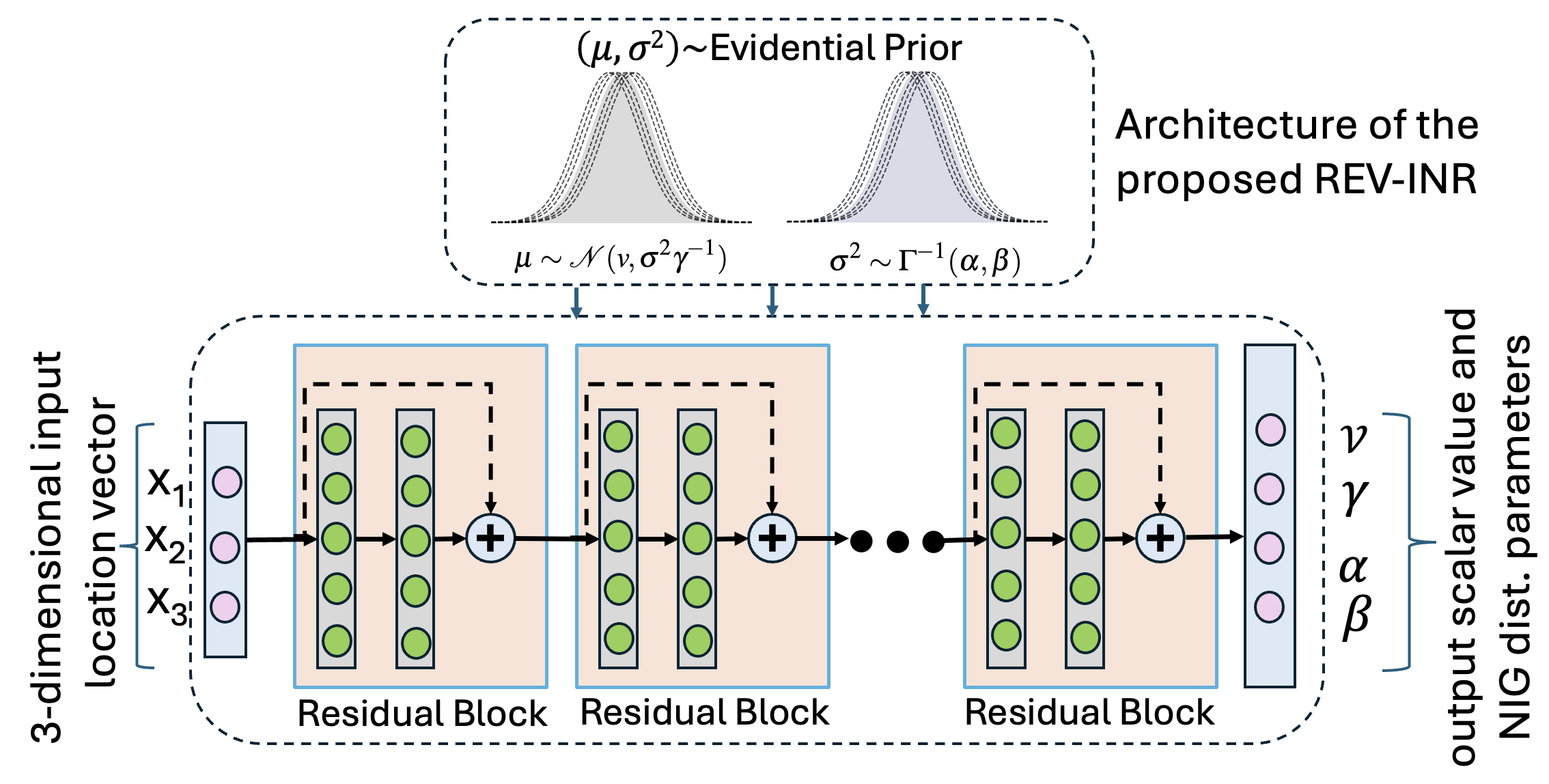}
    \caption{Architecture of the proposed REV-INR.}
    \label{revinr_arch}
\end{subfigure}
~
\begin{subfigure}[t]{0.32\textwidth}
    \centering
    \includegraphics[width=\textwidth]{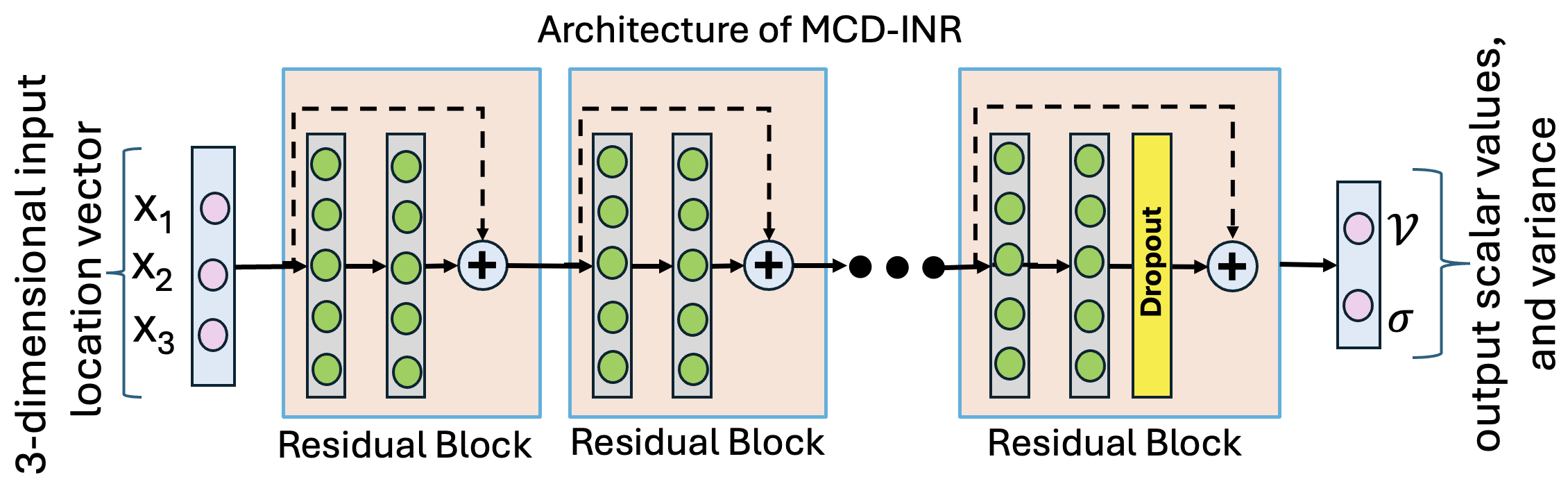}
    \caption{Architecture of MCD-INR.}
    \label{mcdropout_arch}
\end{subfigure}
~
\begin{subfigure}[t]{0.32\textwidth}
    \centering
    \includegraphics[width=\textwidth]{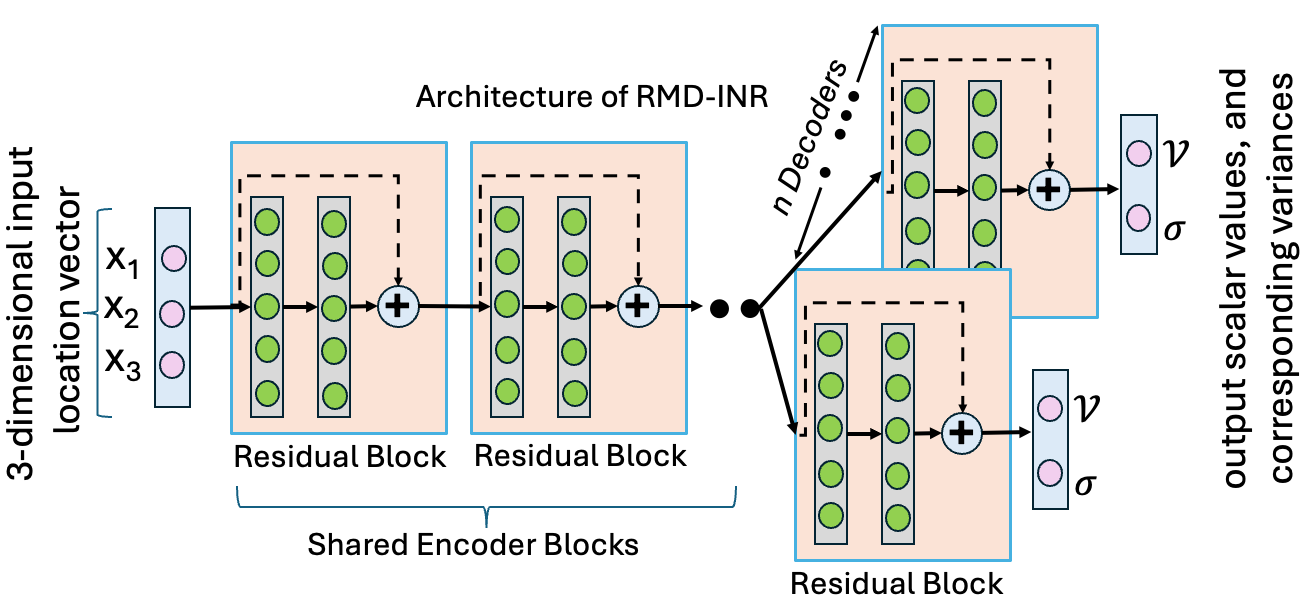}
    \caption{Architecture of RMD-INR.}
    \label{mdinr_arch}
\end{subfigure}
\caption{Schematic architecture of the proposed REV-INR, MCD-INR, and RMD-INR. The proposed REV-INR is trained to predict the parameters of an evidential distribution, modeling a higher-order probability distribution.}
\label{model_arches}
\end{figure*}

\section{Uncertainty in Neural Networks}

Uncertainty refers to the degree of confidence (or lack thereof) in the results, predictions, or visualizations made by a model. It captures the idea that our conclusions from model-generated results are not always absolute but are subject to variability and errors in predictions. In the domain of predictive data analytics, uncertainty can be broadly classified into two types: (1) Epistemic (model) uncertainty, and (2) Aleatoric (data) uncertainty. Studying both types of uncertainty is imperative to generate meaningful, trustworthy, and reliable visual analysis results~\cite{Kendall_UQ}.

\textbf{Epistemic (Model) Uncertainty.}
Epistemic uncertainty, also known as model uncertainty, represents a model's imperfections or limitations regarding its predicted values. It quantifies how confident the model is in the correctness of its predictions and provides a mechanism for users to measure the limitations of the model's knowledge during inference. Recently, the visualization community has started investigating this epistemic uncertainty and its significance in volume visualization tasks~\cite{uncertVolren, Tianyu}.

\textbf{Aleatoric (Data) Uncertainty.}
On the other hand, the importance of Aleatoric (data) uncertainty in volume visualization tasks is yet to be studied. Aleatoric uncertainty captures the inherent variance or noise in the data. It represents the stochastic nature of the data and is often attributed to the limitation of data generation process. Aleatoric uncertainty can be of two types: (1) Homoscedastic and (2) Heteroscedastic~\cite{Kendall_UQ}. In homoscedastic uncertainty, the noise for each input data point is considered constant, whereas the noise in heteroscedastic aleatoric uncertainty varies. Note that, in this work, the estimated aleatoric uncertainties are heteroscedastic since the amount of noise varies across the volume spatially.

\subsection{Uncertainty in INR-based Volume Modeling}
This work proposes using INRs to represent large volumetric datasets in a compact format, enabling downstream analysis and visualization to be performed solely on INR-reconstructed data without requiring access to the ground truth volume. To build trust in INR-predicted data and produce reliable visualization results, we advocate the use of uncertainty-aware INRs over the conventional deterministic INRs. A key advantage of uncertainty-aware INRs is their ability to convey uncertainty information in analysis and visualization results. For example, when visualizing isosurfaces, uncertainties can help recover the true shape and highlight potentially erroneous regions. Similarly, in volume rendering, uncertainty can be quantified and conveyed to domain scientists to support informed decision-making.

Our proposed REV-INR not only predicts data values for given coordinates but also learns its own prediction uncertainty and the inherent noisiness of the volume data. The model’s prediction uncertainty, or epistemic uncertainty, quantifies its confidence in the accuracy of the predicted values. Since the ground truth data is not available, computation of the true error is infeasible. However, the epistemic uncertainty can potentially serve as a proxy for the true error, helping to build trust and improve interpretability. Alongside, the learned aleatoric uncertainty reflects inherent noisiness in the data. This noisiness may arise from (a) stochasticity of the scientific simulation that generates the data, manifested as varying degrees of local variance in the scalar field or (b) from interpolation errors introduced during down-sampling and up-sampling operations of the volume. While down-sampling helps reduce data size, during rendering the data are often up-sampled, and this transformation inevitably introduces interpolation errors. Our goal is for REV-INR’s learned aleatoric uncertainty to correlate with such noise and error sources, allowing their impact on visualization results to be effectively estimated.

\section{Uncertainty-Aware INRs for Reliable Volume Data Modeling and Visualization}

\subsection{Implicit Neural Representation (INR)}
Recent research shows that for coordinate-based datasets, INRs with periodic activation functions efficiently learn the mapping from a given input coordinate to the output data value space~\cite{siren}. The visualization community has proposed several variations of such networks to address several scientific data modeling~\cite{levine_neural_compression, coordnet, stsrinr}. The state-of-the-art results obtained from these deterministic INRs have motivated us to develop REV-INR for efficiently learning volume data representations and  enabling reliable and trustworthy uncertainty-aware volume data visualization, utilizing both aleatoric and epistemic uncertainties. Our base deterministic INR follows a residual SIREN architecture as suggested in~\cite{levine_neural_compression}. The INR takes the 3D coordinates as inputs and outputs the corresponding scalar values. Therefore, the network learns a function $\mathbb{F(\psi)}:\mathbb{R}^3 \mapsto \mathbb{R}$, where $\psi$ denotes the learnable parameters of the INR. 
% Recent research has shown that such INRs are usually significantly smaller in size compared to the raw datasets; hence, these INRs can be conceived as a compact data model for the purpose of achieving compression~\cite{levine_neural_compression}.

\subsection{Proposed Regularized Evidential INR (REV-INR)}\label{sec:revinr}
Construction of a Regularized Evidential Implicit Neural Network (REV-INR) from a base residual INR requires minimal architectural modification, essentially augmenting the final decoder layer with four neurons. The primary novelty lies in the learning algorithm where REV-INR uses an evidential learning scheme~\cite{evidential_survey_arxiv} to estimate data values as well as three additional higher-order distribution parameters, which can then be conveniently used to estimate per-point epistemic and aleatoric uncertainty estimates as a closed-form solution. Therefore, REV-INR has four output neurons at its decoder layer as seen from Fig.~\ref{revinr_arch}. To learn REV-INR, we employ the evidential learning scheme proposed in~\cite{deep_evidential_uncert}. It is assumed that individual data values at grid points are drawn i.i.d. from Gaussian distributions with unknown mean ($\mu$) and variance ($\sigma^2$), which need to be learned. To learn these parameters, an evidential prior is placed on each of these variables, a Gaussian prior for the unknown mean ($\mu$) and Inverse-Gamma prior for the unknown variance ($\sigma^2$). The learning algorithm aims to approximates the true posterior distribution, $z(\mu,\sigma^2)$. Therefor the formulation is as follows: 
\begin{equation}
\label{formulation_eqn}
\mu \sim \mathcal{N}(v, \sigma^2 \gamma^{-1}),~~ \sigma^2 \sim \Gamma^{-1}(\alpha, \beta)
\end{equation}
where $\Gamma()$ is the gamma function, $Q=(v,\gamma,\alpha,\beta)$ and $v\in R$, $\gamma>0$, $\alpha>1$, $\beta>0$. Amini et al. in~\cite{deep_evidential_uncert} show that the approximate posterior distribution comes out to be the Gaussian conjugate prior, which is the Normal Inverse-Gamma (NIG) distribution and the posterior can be expressed using the four parameters $Q=(v,\gamma,\alpha,\beta)$ where the posterior NIG distribution function is represented as:
\begin{equation}
\label{NIG_eqn}
\resizebox{\columnwidth}{!}{$p(\mu, \sigma^2 \mid v, \gamma, \alpha, \beta) = \frac{\beta^\alpha \sqrt{\gamma}}{\Gamma(\alpha) \sqrt{2\pi\sigma^2}}
\left(\frac{1}{\sigma^2}\right)^{\alpha+1} 
\exp \left\{ \frac{-2\beta + \gamma (v - \mu)^2}{2\sigma^2} \right\}$}
\end{equation}
From Equation~\ref{NIG_eqn}, we observe that this NIG distribution has four parameters $Q=(v,\gamma,\alpha,\beta)$. These four parameters are learned per grid point of the volume data so that coordinate-level value and both data and model uncertainty estimation will be possible. 

\subsubsection{Value Prediction and Uncertainty Estimation}
A key benefit of the proposed REV-INR is that, after training, both the estimation of uncertainty values and the prediction of data values can be achieved in a single forward pass, making inference highly efficient and thereby accelerating volume reconstruction. Using the NIG distribution parameters learned by REV-INR, we compute the predicted value and its associated uncertainty estimates at each grid point using the following closed-form expressions~\cite{deep_evidential_uncert}:
\[
\mathbb{E}[\mu] = v, \quad 
AU = \frac{\beta}{\alpha - 1}, \quad 
EU = \frac{\beta}{\gamma(\alpha - 1)} \ 
\]
where $\mathbb{E}[\mu]$ denotes the predicted expected scalar value, $AU$ represents the aleatoric uncertainty ($\mathbb{E}[\sigma^2]$), and $EU$ represents the epistemic uncertainty ($\mathrm{Var}[\mu]$). Note that, $AU$ in the context of volume data, models the inherent data value variance per grid point whereas the $EU$ captures the REV-INR's prediction variability. 

\subsubsection{Loss Functions}
From Fig.~\ref{revinr_arch}, we observe that construction of REV-INR from a base INR requires modification in the last decoder layer to add four neurons so that it can learn to predict the four parameters of the posterior NIG distribution. To learn this evidential distribution parameters, we propose to use a KL-divergence loss along with a regularizer proposed in~\cite{deep_evidential_uncert}. We construct a target NIG distribution which represents low uncertainty when prediction is accurate. Then we minimize the KL-divergence between this target NIG distribution and the predicted NIG distribution by REV-INR. Besides this, we also incorporate a regularization term which applies an
incorrect evidence penalty to minimize evidence on incorrect predictions~\cite{deep_evidential_uncert}. If $\mathcal{L}_{KL}$ denote the KL-divergence loss and $\mathcal{L}_{Reg}$ denotes the evidence regularization term, then the complete evidential loss, $\mathcal{L}_{EV}$, can be written as follows:
\[
\begin{aligned}
\mathcal{L}_{EV} &= \mathcal{L}_{KL} + \delta \cdot \mathcal{L}_{Reg} \\[4pt]
\mathcal{L}_{KL} &= D_{\mathrm{KL}}\!\big(\mathrm{NIG}(v,\gamma,\alpha,\beta) 
\;\|\; \mathrm{NIG}(y,\gamma_t,\alpha_t,\beta_t)\big) \\[4pt]
\mathcal{L}_{Reg} &= \lvert y - v \rvert \cdot (2\gamma + \alpha)
\end{aligned}
\]
where ${NIG}(v,\gamma,\alpha,\beta)$ and  ${NIG}(y,\gamma_t,\alpha_t,\beta_t)$ represent the predicted and target NIG distributions respectively, $y$ denotes the true data value, and $\delta$ is a weight for the evidence regularization term and we set $\delta=0.1$ empirically for all our experiments. Since we want a high reconstruction quality along with meaningful uncertainty estimates, we  incorporate the conventional mean squared error loss ($\mathcal{L}_{MSE}$) with the evidential loss. Hence our loss function becomes:
\[
\begin{aligned}
\mathcal{L} = \mathcal{L}_{MSE} + \lambda_1 \cdot \mathcal{L}_{EV}
\end{aligned}
\]
where $\lambda_1$ is the weight of evidential loss component to achieve a balanced and stable training loss function for REV-INR.

\subsubsection{Regularization of REV-INR}
Training REV-INR using the above loss function ($\mathcal{L}$) results in a stable INR model that yields high reconstruction quality while also enabling per-point epistemic and aleatoric uncertainty quantification. Epistemic uncertainty ($EU$) reflects REV-INR’s confidence in the accuracy of the predicted data, whereas aleatoric uncertainty ($AU$) captures the inherent data noisiness. Both types of uncertainty can play a crucial role when visualization results are generated using REV-INR predictions, as incorporating uncertainty information allows domain experts to better interpret the results and make informed and reliable decisions. However, as prior research has shown~\cite{Tianyu}, such predicted uncertainty values—representing either model variance or data variance—may be over- or under-confident, potentially leading to misleading interpretations. Therefore, proper calibration is necessary to produce interpretable uncertainty estimates. We address this challenge by formulating novel regularization methods for both $EU$ and $AU$. Unlike typical post-training calibration approaches, where predicted uncertainty values are adjusted to align with prediction error~\cite{Tianyu}, we perform calibration during training itself. Specifically, we introduce novel regularization terms into our loss function that directly constrain $EU$ and $AU$, thereby yielding interpretable and reliable uncertainty estimates.

\textbf{Epistemic Uncertainty Regularization.}
Since $EU$ represents REV-INR’s confidence in the accuracy of the predicted data values, it is expected to correlate with the prediction error. However, because visual analysis relies solely on REV-INR–generated data, where ground-truth values are unavailable, direct estimation of the true error is infeasible. In this context, $EU$ can serve as a potential proxy for the error values~\cite{Tianyu}. Therefore, it is appropriate to regularize the raw $EU$ values against the prediction error during training. We achieve this by introducing a correlation loss component between the predicted $EU$ values and the corresponding true error values. This is feasible because, during training, we have access to the raw data and can compute the actual error. Hence, our $EU$ regularization loss ($\mathcal{L}_{EU}$) is defined as follows:
\[
\begin{aligned}
\mathcal{L}_{EU} = 1 - \rho(EU,\xi)
\end{aligned}
\]
where $\rho(EU,\xi)$ denotes the Pearson correlation between the prediction error ($\xi$) and $EU$. Note that when $EU$ and $\xi$ are positively correlated, $\mathcal{L}_{EU}$ is low, whereas if they are negatively correlated, $\mathcal{L}_{EU}$ yields higher loss. This loss component encourages REV-INR to produce $EU$ that are aligned with the prediction error.

\textbf{Aleatoric Uncertainty Regularization.}
$AU$ captures the inherent noisiness of the volume at each grid point. In volumetric datasets, such noise can arise from different sources and may be interpreted in multiple ways. During training, the proposed REV-INR learns a continuous representation of the scalar function so that, given any 3D coordinate as input, it can predict the corresponding scalar value. While learning this scalar function, REV-INR also models per-point noise in the form of $AU$. Consequently, homogeneous regions of the volume are expected to contain less noise compared to regions where scalar values change rapidly or where edges exist. Therefore, $AU$ should faithfully represent such variations. Identification of homogeneous regions or rapidly changing regions can be achieved by computing per-grid-point gradient values within a small local neighborhood. Gradient magnitude is an inherent property of the data and serves as an indicator of regions where high $AU$ is likely to coincide. To regularize REV-INR-predicted $AU$, we therefore correlate $AU$ values with local gradient magnitudes. Our hypothesis is that this regularization leads to more meaningful $AU$ estimates, which in turn can serve as a reliable proxy for the inherent noisiness of volumetric data. To demonstrate the validity and usefulness of the learned $AU$ values, we later show that these regularized $AU$ estimates exhibit high correlation with both the local data variance and interpolation-based inaccuracies that can arise when volume data are down-sampled and subsequently up-sampled for rendering. Hence, the $AU$ regularization loss ($\mathcal{L}_{AU}$) is defined as follows:
\[
\begin{aligned}
\mathcal{L}_{AU} = 1 - \rho(AU,\nabla f(\mathbf{\cdot}))
\end{aligned}
\]
where $\rho(AU,\nabla f(\mathbf{\cdot}))$ denotes the Pearson correlation between the gradient magnitude ($\nabla f(\mathbf{\cdot})$) and $AU$, with $f(\cdot)$ being the scalar function. Note that if $AU$ and $\nabla f(\mathbf{\cdot})$ are positively correlated, $\mathcal{L}_{AU}$ is low, whereas if $AU$ and $\nabla f(\mathbf{\cdot})$ are negatively correlated, $\mathcal{L}_{AU}$ is high.

\subsubsection{Training REV-INR with Combined Loss Functions}
The proposed regularized evidential INR model (REV-INR) is trained using multiple loss functions combined to achieve both high reconstruction quality and meaningful, reliable uncertainty estimates. We observe that training REV-INR with all loss components throughout all epochs can slightly reduce reconstruction quality and increase training time significantly. To alleviate this, we adopt a two-phase training strategy. In the first phase, we train REV-INR using only $\mathcal{L}_{MSE}$, bringing the model into a stable state with high prediction quality. In the second phase, we introduce $\mathcal{L}_{EV}$ along with the two uncertainty regularization terms, $\mathcal{L}_{AU}$ and $\mathcal{L}_{EU}$, and continue training for the remaining epochs. This strategy yields a stable and reliable REV-INR model that achieves superior prediction quality while producing meaningful and calibrated uncertainty estimates. Therefore, if REV-INR is trained for a total of $n$ epochs, the final training strategy is defined as follows:
\[
\begin{gathered}
\mathcal{L}_{REV-INR} = \mathcal{L}_{MSE}, \quad \text{for } 1 < \text{epoch} < n/2 \\[6pt]
\resizebox{\columnwidth}{!}{$\mathcal{L}_{REV-INR} = \mathcal{L}_{MSE} + \lambda_1 \mathcal{L}_{EV} + \lambda_2 \mathcal{L}_{EU} + \lambda_3 \mathcal{L}_{AU}, 
\quad \text{for } n/2 \leq \text{epoch} \leq n$}
\end{gathered}
\]
where $\lambda_1$, $\lambda_2$, and $\lambda_3$ are three weight parameters used to balance the influence of evidential loss and the two uncertainty regularization components during training. These weights are chosen empirically and more information can be found in the supplementary material. 

\subsection{Monte Carlo Dropout INR (MCD-INR) Enhanced with Aleatoric Uncertainty Estimation Capability}\label{sec:mcdropout}
To build a Monte Carlo Dropout–based INR (MCD-INR), we first insert dropout layers into the base residual INR model. Kendall et al.~\cite{segnet} and Saklani et al.~\cite{uncertVolren} demonstrate that applying dropout after every layer can act as a strong regularizer, often degrading prediction quality. Therefore, a dropout layer after only a subset of layers—or solely after the last layer—can be used to simulate a partial Bayesian neural network. Following this strategy, we apply post-activation dropout only at the last residual block, which allows us to obtain high-quality predictions~\cite{uncertVolren}. With this design, MCD-INR predicts only data values and can estimate $EU$ by activating dropout during inference and computing the variance in predictions across multiple forward passes. In this work, we extend conventional MCD-INR to also estimate $AU$, enabling the model to learn the inherent noisiness of volumetric data at each grid point. To achieve this, we add an additional neuron in the output decoder layer so that the model simultaneously predicts the scalar value and its associated per-point variance. A standard way to jointly learn mean and variance is to use the Gaussian negative log-likelihood (NLL) as the loss function. Thus, MCD-INR is trained to output both the predicted mean and variance for each grid point. The overall MCD-INR architecture is illustrated in Fig.~\ref{mcdropout_arch}.

\subsubsection{Value Prediction and Uncertainty Estimation}
During inference we first enable dropout. For a given input coordinate, $n$ predictions are obtained by performing $n$ stochastic forward passes through the model. The average of these $n$ predictions is taken as the expected scalar value, while the average of the $n$ predicted variances represents the $AU$. Finally, the variance of the $n$ predicted scalar values across the forward passes provides the $EU$. Hence, considering $n$ stochastic forward passes for an input, if $\hat{y}_i$ denotes the predicted scalar value and $\sigma_i^2$ the corresponding variance for the $i$-th forward pass, the following expressions are used to compute the expected scalar value ($\mu$), $EU$, and $AU$:
\[
\begin{aligned}
\mathbb{E}[\mu] = \frac{1}{n} \sum_{i=0}^{n} \hat{y}_i, \quad 
AU = \frac{1}{n} \sum_{i=0}^{n} \sigma_i^2, \quad 
EU = Var(\hat{y}_i) \, 
\end{aligned}
\] 

\subsubsection{Loss Functions} \label{mcd_loss_func}
To simultaneously learn the values and their associated variances, the Gaussian negative log-likelihood (NLL) is used as the loss function. The Gaussian NLL loss function is given by:
\[
\begin{aligned}
\mathcal{L}_{\text{GaussNLL}} 
= \frac{1}{2} \log \big(2\pi\sigma^2\big) 
+ \frac{(y - \mu)^2}{2\sigma^2}
\end{aligned}
\] 
Following a strategy similar to REV-INR, we incorporate the conventional mean squared error loss to achieve both high reconstruction quality and reliable uncertainty estimates. The final loss function used to train the MCD-INR, with  $\lambda_1$ being the weight for the $\mathcal{L}_{\text{GaussNLL}}$, is therefore given by:
\[
\begin{aligned}
\mathcal{L}_{MCD-INR} = \mathcal{L}_{MSE} + \lambda_1 \cdot \mathcal{L}_{\text{GaussNLL}} 
\end{aligned}
\]

\subsection{Regularized Multi-Decoder INR  (RMD-INR) Equipped with Aleatoric Uncertainty Estimation Capability}\label{sec:mdinr}
We adopt from the shared-encoder, multi-decoder architecture proposed in~\cite{Tianyu} to build a the  Regularized Multi-Decoder INR (RMD-INR). In our implementation, we use MLPs for both the encoder and decoders of RMD-INR. We call this new model RMD-INR, and enhance it with the ability of $AU$ estimation, which was lacking in RMDSRN. The RMD-INR consists of five decoder heads with each decoder independently predicting the scalar value. To enable each decoder head to predict both the scalar value and its associated $AU$, we add an additional neuron at the output layer of each decoder. This modified model architecture is illustrated in Fig.~\ref{mdinr_arch}.

\subsubsection{Value Prediction and Uncertainty Estimation}
During inference, each decoder head of the RMD-INR predicts both the scalar value and the associated $AU$. The final expected scalar value is obtained by averaging the scalar values predicted by all decoder heads. Similarly, $AU$ is estimated by averaging the $AU$ values predicted by the decoder heads, while the variance of the scalar values represents the $EU$. Hence, considering $D$ decoder heads, if $\hat{y}_i$ denotes the predicted scalar value and $\sigma_i^2$ the corresponding variance from the $i$-th decoder head, the following expressions can be used to compute the expected scalar value ($\mu$), $AU$, and $EU$:
\[
\begin{aligned}
\mathbb{E}[\mu] = \frac{1}{D} \sum_{i=0}^{D} \hat{y}_i, \quad 
AU = \frac{1}{D} \sum_{i=0}^{D} \sigma_i^2, \quad 
EU = Var(\hat{y}_i) \, 
\end{aligned}
\] 

\subsubsection{Loss Functions}
RMD-INR is trained using the same loss functions as MCD-INR, discussed above in~\ref{mcd_loss_func}. To enforce meaningful $EU$ estimation, we follow the regularization strategy suggested in~\cite{Tianyu} for RMDSRN. Hence, we incorporate a KL-Divergence regularization term in the RMD-INR's loss function which enforces minimization between predicted variance ($EU$) and prediction error values. We follow the exponential growth weight scheduler as proposed in~\cite{Tianyu} to vary the weight of the regularizer term during training. Therefor the final loss function is given as follows:
\[
\begin{aligned}
\mathcal{L}_{RMD-INR} = \mathcal{L}_{MSE} + \lambda_1 \cdot \mathcal{L}_{\text{GaussNLL}} + \lambda_2 \cdot D_{KL}(EU,error)
\end{aligned}
\] 
where $\lambda_1$ is the weight for the $\mathcal{L}_{\text{GaussNLL}}$ loss component and $\lambda_2$ is the weight for the KL-divergence, $D_{KL}(EU,error)$.

\subsection{Hyperparameters}
The base deterministic INR consists of a residual block-based architecture, where each residual block has two layers with $100$ neurons each. Following the suggestions reported in~\cite{kdinr}, we make our INRs wider rather than deeper to achieve a better storage–quality trade-off. We use three residual blocks to construct REV-INR and MCD-INR. We design RMD-INR to occupy the same disk space for storing model parameters as MCD-INR and REV-INR, in order to compare all methods under a similar number of parameters and storage requirements. Hence, for RMD-INR, each layer consists of $70$ neurons, and there are five decoder heads (see Fig.\ref{mdinr_arch}), as suggested in\cite{Tianyu}. For training all models, including our REV-INR across all datasets, we conduct empirical experiments to select a suitable and consistent learning rate and batch size that produce stable and high-quality predictions. We employ the Adam optimizer~\cite{kiba14} with a learning rate of $0.00005$, and the two optimizer coefficients $\beta_1$ and $\beta_2$ are set to $0.9$ and $0.999$, respectively.  For all methods, a learning rate decay scheme is applied with a decay factor of $0.8$ and a step size of $15$. All models are trained for up to $300$ epochs to ensure convergence. For all datasets, the weight of the Gaussian NLL loss component is set to $0.001$ for both MCD-INR and RMD-INR. 

\section{Results}
\label{results}

\begin{figure}[!t]
\centering 
\includegraphics[width=\linewidth]{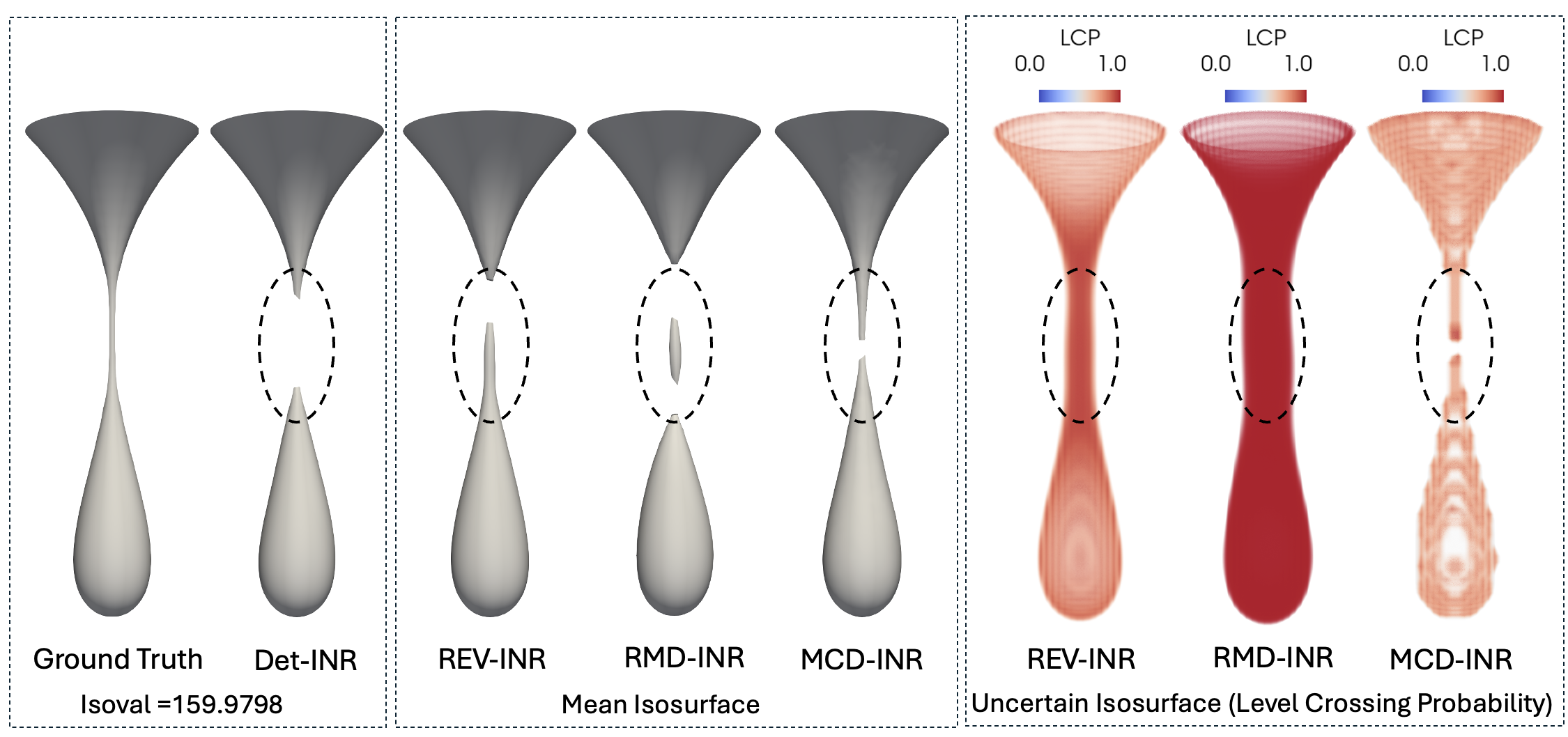}
\caption{Visualization of the Teardrop Dataset for isovalue $159.9798$. The middle segment shows the mean isosurfaces generated by REV-INR, RMD-INR, and MCD-INR, respectively, where differences from the ground truth are highlighted (dashed circles). The right segment shows the corresponding uncertain isosurfaces using the Level Crossing Probability (LCP). We observe that REV-INR produce the most accurate LCP visualization.}
\label{teardrop_isocontour}
\end{figure}

\begin{figure*}[!t]
\centering 
\includegraphics[width=\linewidth]{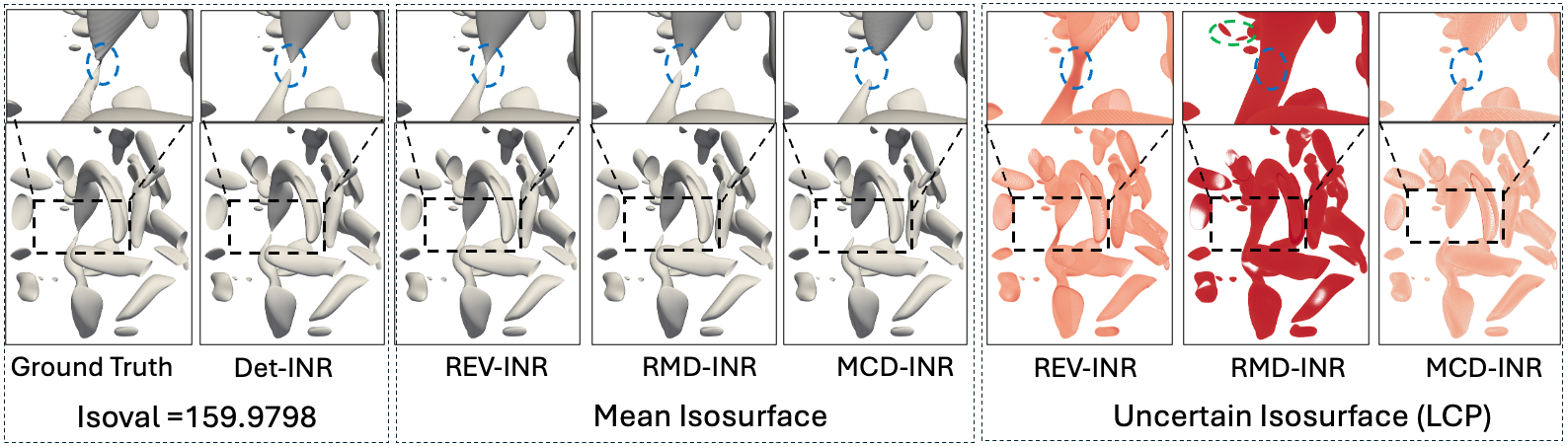}
\caption{Visualization of the Vortex Dataset for the isovalue 5.8. The middle column shows the mean isosurfaces generated by REV-INR, RMD-INR, and MCD-INR, respectively, where differences from the ground truth are highlighted(dashed circles). The right column shows the corresponding uncertain isosurfaces using the Level Crossing Probability (LCP), showing the most probable isosurface. We observe that REV-INR produce the most accurate LCP visualization. The isosurface generated by Det-INR also fails to preserve the thin connection.}
\label{vortex_isocontour}
\end{figure*}

In this section we present qualitative and quantitative results. The spatial resolution, the size of each INR, raw data  are reported in Table~\ref{inr_results}. We use a GPU server with NVIDIA RTX A$6000$ GPUs, each with $48$ GB of GPU memory, for experimentation.

\begin{figure*}[!t]
\centering 
\includegraphics[width=0.8\linewidth]{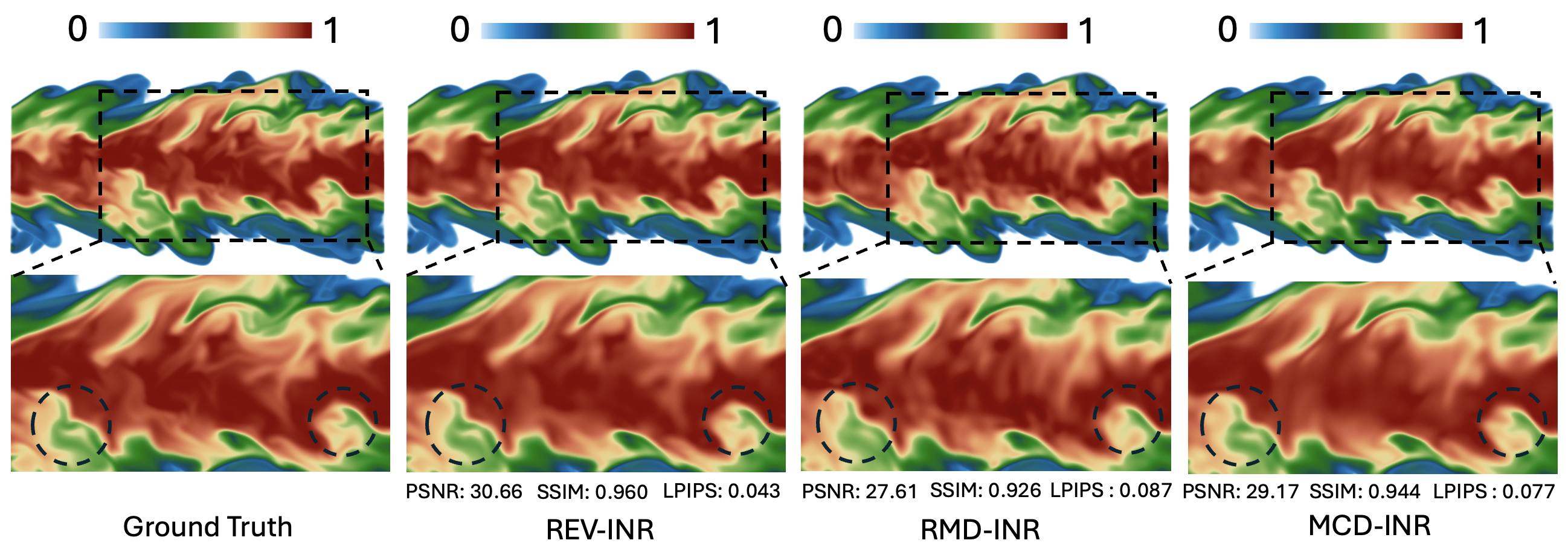}
\caption{Visualization of the Combustion Dataset. The ground truth scalar field is compared with REV-INR, RMD-INR, and MCD-INR. REV-INR better preserves complex flame structures, as highlighted in the zoomed-in regions marked with black boxes.}
\label{comb_volren}
\end{figure*}

\begin{figure}[!t]
\centering 
\includegraphics[width=\linewidth]{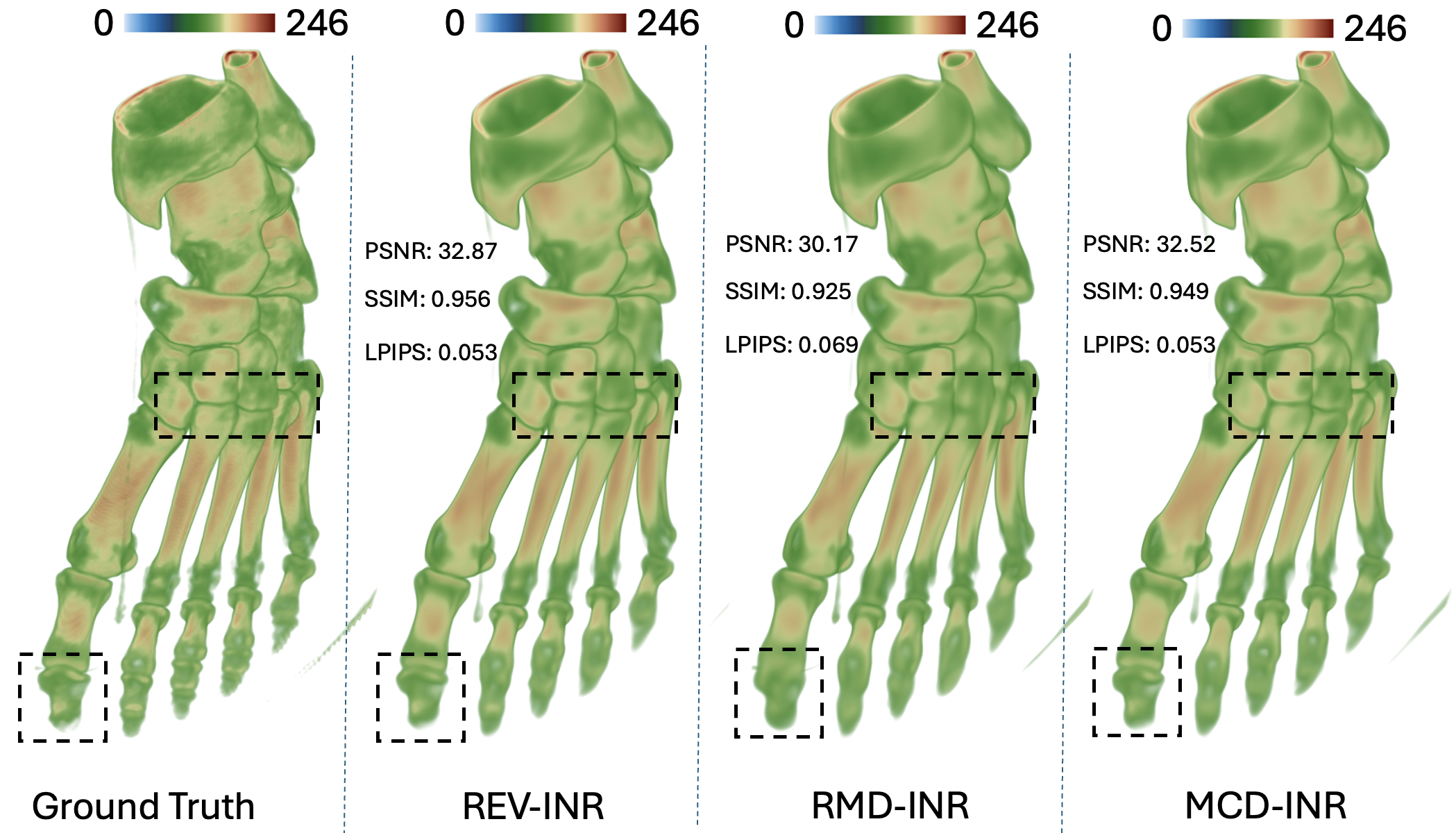}
\caption{Visualization of the Foot Dataset. The ground truth scalar field (left) is compared with reconstructions from REV-INR, RMD-INR, and MCD-INR. The highlighted regions around the joints of bone segments indicate that REV-INR achieves the closest match to the ground truth compared to the other methods.}
\label{foot_volren}
\end{figure}

\subsection{Uncertainty-Informed Isosurface Visualization} \label{uncert_isosurf}
We demonstrate how uncertainty estimates can help improve isosurface extraction and visualization, making them more reliable and informative. We show that the isosurfaces extracted using only the predicted mean fields contain inaccuracies. In contrast, uncertainty information can help identify under- or over-confidently predicted isosurface regions, allowing users to locate potential errors. Since our uncertainty-aware INRs produce uncertainty estimates at each grid location, we utilize isosurface uncertainty quantification and visualization techniques to generate informative and trustworthy results. This approach also helps validate that the uncertainty estimates produced by REV-INR are meaningful and essential for creating uncertainty-informed visualizations. Consequently, we employ uncertainty visualization techniques on the isosurfaces extracted from INR-reconstructed scalar fields, which are pivotal for making INR predictions more reliable and trustworthy by effectively communicating uncertainty. For completeness and comparative purposes, we also include results from deterministic INRs (Det-INR). From Table~\ref{inr_results}, we observe that while the Det-INRs achieve slightly higher or comparable reconstruction PSNR, they still produce erroneous isosurfaces, which can cause inaccurate feature interpretation. Hence, uncertainty-aware INRs are essential to remedy such issues.

Since uncertain INRs produce distribution-based volume, we employ uncertain isosurface algorithms~\cite{Wang2023, Athawale_UncertaintyMarchingCubes} for producing uncertainty-informed visualizations. We visualize the \textit{level-crossing probability (LCP)} within a cell. The LCP implicitly communicates uncertainty by computing the probability of an isosurface crossing through a grid cell. Thus, high LCP values indicate a higher likelihood of isosurface presence. Our uncertainty-aware INRs generate a predicted mean $\mu$ and  variance $\sigma^2$ at each grid location. Therefore, we use the parametric Gaussian distribution-based uncertainty modeling for isosurface extraction.

\textbf{Teardrop Dataset.} We present results for uncertainty-informed isosurface visualization of the Teardrop dataset~\cite{teardropdata} in Fig.~\ref{teardrop_isocontour} for the isovalue $159.9798$. To generate uncertain isosurfaces, we use EU values as the variance at each grid location, while the predicted scalar values are treated as the mean of the Gaussian distribution. We observe that the predicted mean isosurfaces for all three methods are broken at the central region when compared to the ground truth. The isosurface generated by Det-INR also produce incorrect isosurface. This indicates that the deterministically extracting isosurfaces using only the mean values or using a deterministic INR can be erroneous. Examining the LCP field for the three methods, we find that both REV-INR and RMD-INR recover the thin connecting region with high probability, but RMD-INR tends to overestimate the structure of the connection, resulting in a larger spatial spread compared to REV-INR. In contrast, LCP of MCD-INR fails to recover the connection altogether. These results highlight the superiority of REV-INR over MCD-INR, RMD-INR, and Det-INR, reinforcing the advantages of uncertainty-aware INRs and demonstrating how uncertainty information can lead to more reliable isosurface visualization.

\textbf{Vortex Dataset.} In Fig.~\ref{vortex_isocontour}, we present results of mean and uncertain isosurface visualization for the Vortex dataset~\cite{vortex_dataset} at isovalue $5.8$. While exploring the vortex features, an important task is to correctly detect the merge and split events. Keeping this task in mind, from the mean isosurfaces, we observe that both RMD-INR and MCD-INR miss the connection between two tubular vortices, as highlighted by the blue circular regions in the zoomed figure. Isosurface generated by Det-INR also fails to preserve this,  resulting inaccurate feature interpretation. The LCP fields reveal that while REV-INR and RMD-INR recover the thin connecting region, RMD-INR spatially overestimates the isosurface. The LCP of MCD-INR fails to recover the connection. We also highlight another region in the LCP of RMD-INR showing inaccuracy in LCP. In contrast, REV-INR provides a crisp and accurate representation of the isosurface, demonstrating REV-INR's superiority, even over the Det-INR with a slightly higher overall reconstruction PSNR.

\subsection{Reconstructed Volume Visualization} \label{volume_vis}
First, we perform direct volume visualization using the reconstructed scalar fields to demonstrate the superiority of REV-INR over MCD-INR and RMD-INR. From Table~\ref{inr_results}, we observe that REV-INR achieves the best reconstruction quality, among the uncertainty-INRs, in terms of PSNR. To further validate this, we conduct a qualitative visual analysis through volume rendering. In Fig.\ref{comb_volren} and Fig.\ref{foot_volren}, we present the volume rendering results for the Combustion and Foot datasets, respectively. For all methods, we use a consistent transfer function setup and color scale to ensure fair comparisons. From Fig.\ref{comb_volren}, we observe that several regions containing complex flame structures are better preserved by REV-INR compared to MCD-INR and RMD-INR. The zoomed-in regions, highlighted with black boxes, illustrate these differences. Similarly, in Fig.\ref{foot_volren}, REV-INR produces the most accurate volume visualizations for the Foot dataset when compared to the other two methods. The highlighted regions around the joints of different bone segments in Fig.~\ref{foot_volren} clearly show that REV-INR achieves the closest match to the ground truth. We further compute image-level metrics—PSNR, SSIM, and LPIPS—and report their values in Fig.\ref{comb_volren} and Fig.\ref{foot_volren} for all three methods. We observe that REV-INR outperforms MCD-INR and RMD-INR across all three metrics, producing the most accurate volume visualizations.

\begin{figure}[!t]
\centering 
\includegraphics[width=\linewidth]{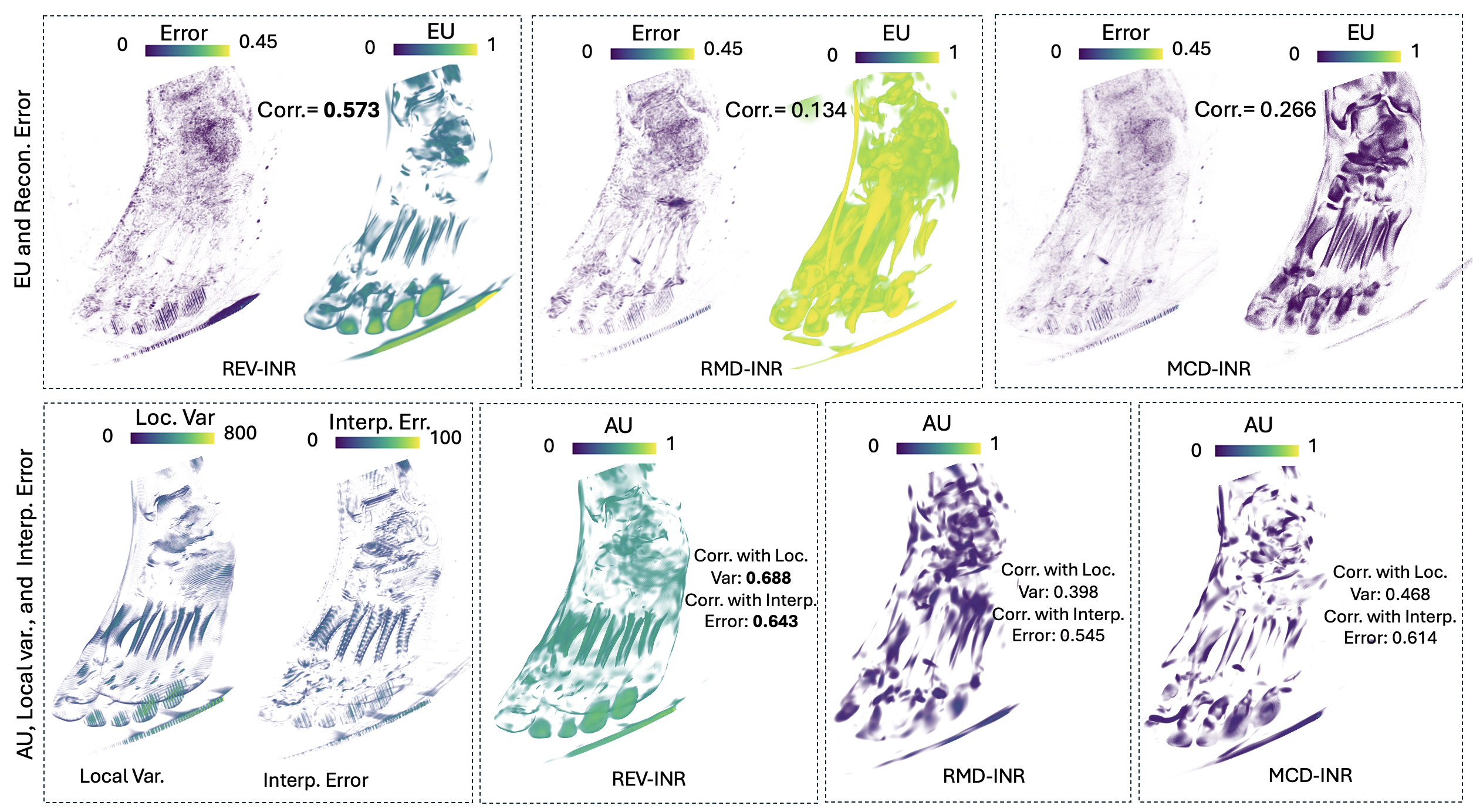}
\caption{\rmark{Volume visualizations of EU and error fields for REV-INR, RMD-INR, and MCD-INR for the Foot dataset (top). REV-INR’s EU aligns closely with high-error regions, while RMD-INR and MCD-INR show overconfident predictions in low-error areas. The bottom row shows visualization of local variance, interpolation error, and AU. REV-INR’s AU field aligns best with both local variance and interpolation error, achieving the highest correlations.}}
\label{foot_uncert_vis}
\end{figure}

\begin{figure}[!t]
\centering 
\includegraphics[width=\linewidth]{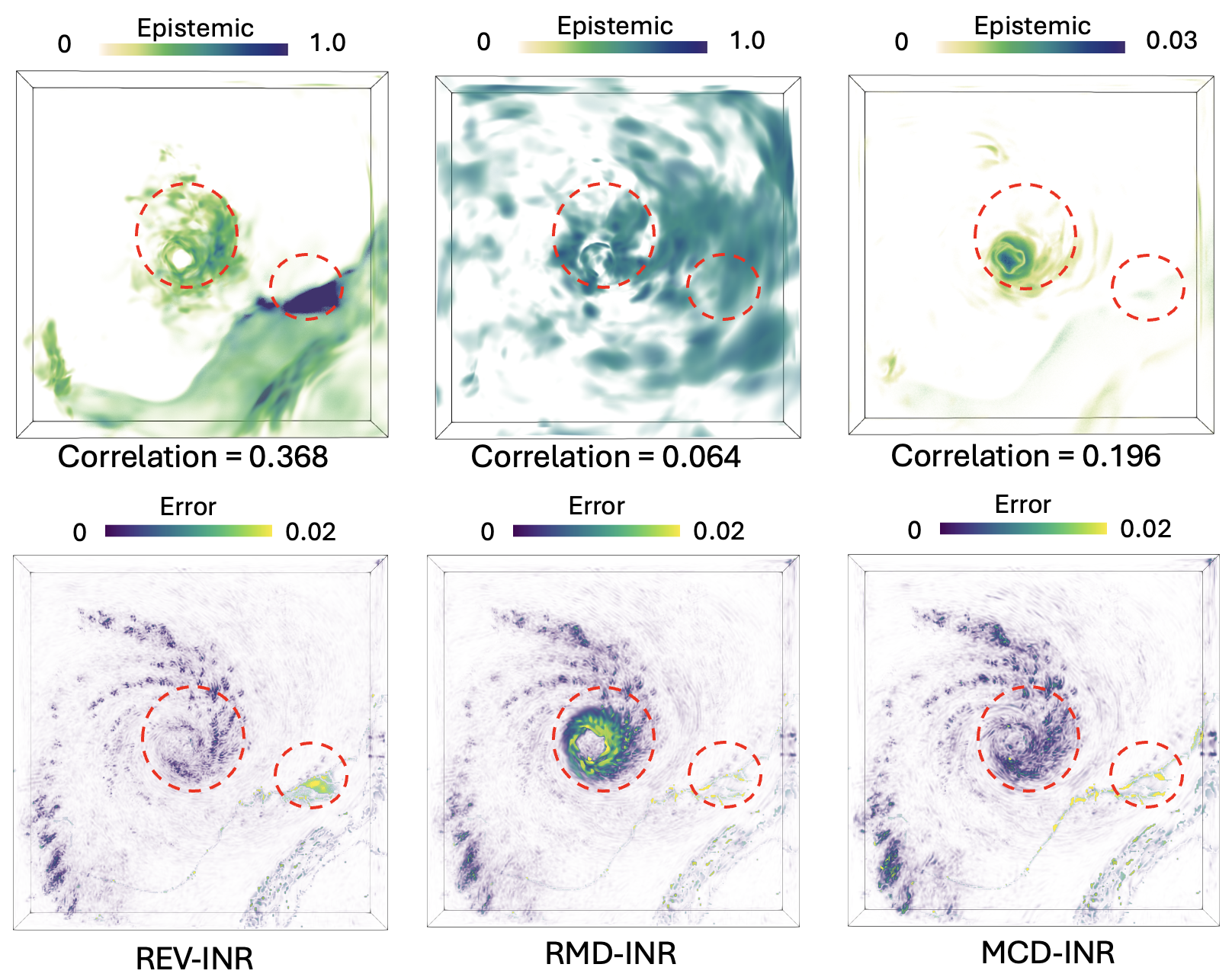}
\caption{\rmark{Visualization of EU and error fields generated by REV-INR, RMD-INR, and MCD-INR for the velocity field of the Hurricane Isabel dataset. REV-INR’s EU aligns best with error fields, with correlation values confirming REV-INR’s superiority.}}
\label{isabel_vel_epistemic_vis}
\end{figure}

\begin{figure}[!t]
\centering 
\includegraphics[width=\linewidth]{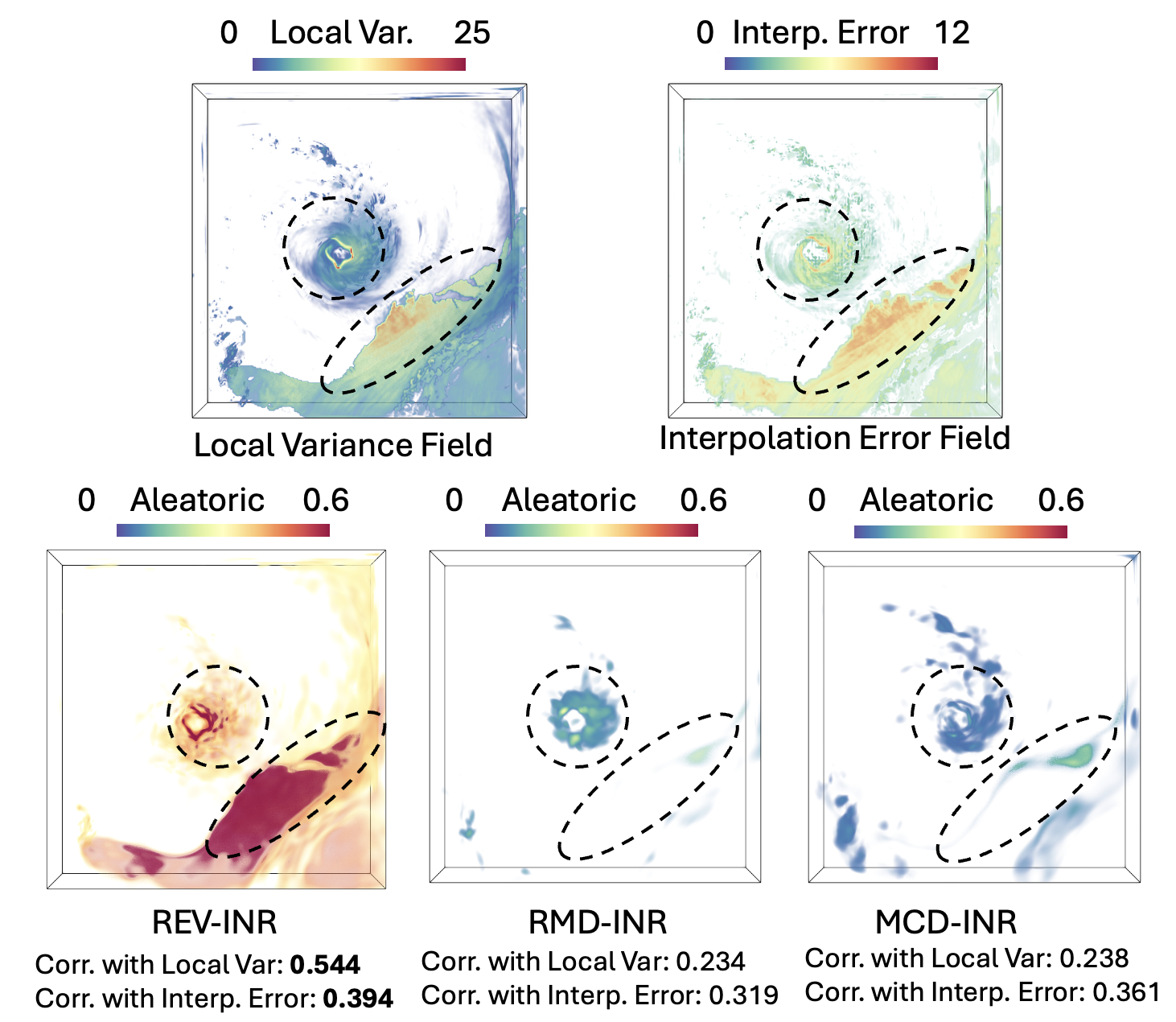}
\caption{\rmark{Visualization of AU, local variance, and interpolation error for the velocity field of the  Isabel dataset. REV-INR shows the highest spatial overlap and correlation, indicating more reliable AU estimates than RMD-INR and MCD-INR.}}
\label{isabel_vel_aleatoric_vis}
\end{figure}

\subsection{Evaluation of Uncertainty Estimates}

\subsubsection{Evaluation of Epistemic Uncertainty (EU) Estimates}
Assessment of the quality of uncertainty estimates is important for understanding their usability. Epistemic uncertainty reflects the model's prediction uncertainty, reflecting the model's confidence in the accuracy of the predicted values. It is an inherent property of the model. Since we deal with large volumetric data and advocate for downstream analysis tasks to be solely driven by model-predicted data, we recognize that raw data may not be available for error quantification during visual analysis. In such scenarios, EU can serve as a proxy for prediction error and can be used to produce interpretable and informative visualizations that highlight potentially erroneous regions for domain experts. Here, we present comparisons between volume visualizations of EU and the corresponding error fields for REV-INR, RMD-INR, and MCD-INR to compare spatial correlations between EU and error fields across the three method. A higher spatial correlation indicates more interpretable EU estimation.

The top half of Fig.\ref{foot_uncert_vis} shows EU and error field visualizations for the Foot dataset. It is observed that the EU field produced by REV-INR shows higher and meaningful spatial overlap with highly erroneous regions when compared against the EU and error fields generated by RMD-INR and MCD-INR. The EU fields of RMD-INR and MCD-INR contain some high EU regions where the error is low, indicating potential overconfident predictions from RMD-INR and MCD-INR. In comparison, the EU predicted by REV-INR aligns well with the high-error regions. We also report the quantitative correlation values in Fig.\ref{foot_uncert_vis}, which also show higher correlation for REV-INR. A similar analysis using the velocity field of the Hurricane Isabel dataset is shown in Fig.\ref{isabel_vel_epistemic_vis}. We observe a similar trend, where the EU and error fields generated by REV-INR are better aligned compared to RMD-INR and MCD-INR. The EU fields generated by RMD-INR show poor spatial correlation, as illustrated in Fig.\ref{isabel_vel_epistemic_vis}. The reported correlation values also support these observations, showing that the EU estimates produced by REV-INR serve as a more appropriate proxy for prediction errors and can be used to produce reliable and informative visualization results.

\subsubsection{Evaluation of Aleatoric Uncertainty (AU)}
Aleatoric uncertainty (AU) captures the noisiness in the data. In volumetric data, such noise can arise from the stochasticity of the simulation and local neighborhood-based variance values can serve as an indicator of that. To reduce data size, volume data is also often down-sampled and later, during rendering, interpolated back to the original resolution to achieve high-quality visualization. This process of down-sampling and subsequent up-sampling introduces interpolation errors, which we consider another potential source of noise. Since AU captures data uncertainty, it is expected to reflect these types of noise. While regularizing REV-INR, we hypothesized that regularizing AU values during training using gradient magnitudes can help produce meaningful and interpretable AU values. The assumption is that local gradient magnitudes can serve as potential cues to identify regions where high noise is likely to be introduced during data transformations, such as  interpolation–based up/down-sampling. Gradient magnitudes are also conceptually expected to show correlation with local data variance values. To study the estimated AU quality, we compute correlations between (1) AU and local variance and (2) AU and interpolation error.

The bottom half of Fig.~\ref{foot_uncert_vis} shows results for the Foot dataset. We apply an 8$\times$8$\times$8 down-sampling to compute the interpolation error field and a 2$\times$2$\times$2 window for local variance estimation. The AU field produced by REV-INR exhibits the highest spatial overlap with both the local variance and interpolation error fields compared to RMD-INR and MCD-INR. Consistently, the correlations between (1) AU and local variance and (2) AU and interpolation error are also highest for REV-INR, as reported in Fig.~\ref{foot_uncert_vis}. A similar analysis on the velocity field of the Isabel dataset is shown in Fig.~\ref{isabel_vel_aleatoric_vis}, where a 5$\times$5$\times$5 down-sampling and a 2$\times$2$\times$2 window are used. Visual comparison indicates that REV-INR achieves the strongest overlap with both local variance and interpolation error among all methods. The higher correlation values further confirm that REV-INR provides superior AU estimates compared to RMD-INR and MCD-INR.

\subsection{Comparison with Compression Methods} \label{compress_compare}
\begin{figure*}[!t]
\centering 
\includegraphics[width=0.8\linewidth]{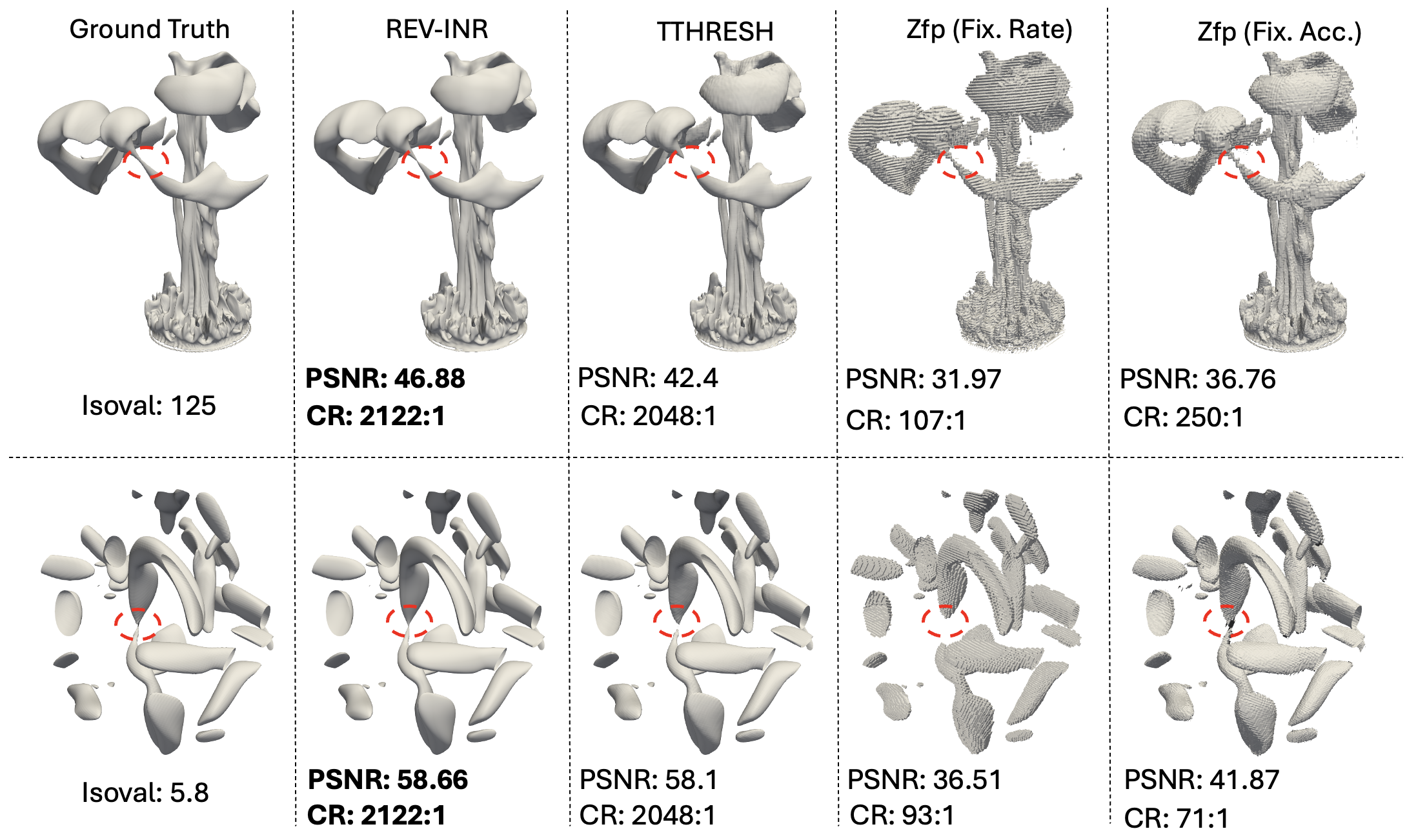}
\caption{Comparison of REV-INR with TTHRESH and Zfp. It is observed that REV-INR achieves the best compression ratio (CR) vs. PSNR trade-off as well as the most accurate isosurface visualization.}
\label{compression_compare}
\end{figure*}
In Fig.~\ref{compression_compare}, we compare the proposed REV-INR with two state-of-the-art compression methods: TTHRESH~\cite{TTHRESH} and Zfp~\cite{zfp}. For Zfp, we perform compression under both fixed-accuracy and fixed-bitrate settings. The comparison is conducted using the Vortex and Heptane datasets. From Fig.~\ref{compression_compare}, we observe that REV-INR produces the most accurate isosurface visualizations, while TTHRESH and Zfp fail to preserve intricate isosurface features (highlighted by the red dotted regions). We also observe that Zfp introduces noticeable visual artifacts in the isosurfaces. In Fig.~\ref{compression_compare}, we further provide the reconstruction PSNR and compression ratio for quantitative comparison, showing that REV-INR achieves the best compression ratio (CR) vs. PSNR trade-off compared to TTHRESH and Zfp.

\section{Quantitative Evaluation}
\label{quant_study}
\begin{table*}[!t]
\centering
\caption{Quantitative evaluation of REV-INR, MCD-INR, and RMD-INR using various performance metrics across multiple datasets. In PSNR and Recon. Time columns, the best values are boldfaced and underlined, while the second-best values are boldfaced only.}
\label{inr_results}
\resizebox{\textwidth}{!}{%
\begin{tabular}{|c|c|c|c|c|c|c|c|c|c|c|}
\hline
\textbf{Dataset} &
  \textbf{Model} &
  \textbf{\begin{tabular}[c]{@{}c@{}}Model \\ Size (KB)\end{tabular}} &
  \textbf{PSNR (dB) $\uparrow$} &
  \textbf{\begin{tabular}[c]{@{}c@{}}Corr. (EU \\ and Error) $\uparrow$ \end{tabular}} &
  \textbf{\begin{tabular}[c]{@{}c@{}}Corr. (AU and \\ Loc. Var.) $\uparrow$ \end{tabular}} &
  \textbf{\begin{tabular}[c]{@{}c@{}}Corr. (AU and \\ Interp. Err.) $\uparrow$ \end{tabular}} &
  \textbf{NLL EU $\downarrow$} &
  \textbf{NLL AU $\downarrow$} &
  \textbf{Train Time (Hrs.)} &
  \textbf{\begin{tabular}[c]{@{}c@{}}Recon.\\  Time (Secs.) $\downarrow$\end{tabular}} \\ \hline
\multirow{4}{*}{\begin{tabular}[c]{@{}c@{}}Teardrop\\ 128x128x128\\ 8MB\end{tabular}} &
  REV-INR &
  246 &
  \underline{\textbf{78.06}} &
  \textbf{0.445} &
  \textbf{0.889} &
  \textbf{0.73} &
  -0.973 &
  -0.35 &
  1.16 &
  \textbf{0.13} \\ \cline{2-11} 
 &
  MCD-INR &
  246 &
  75.91 &
  0.234 &
  0.086 &
  0.034 &
  \textbf{-6.17} &
  -5.95 &
  0.43 &
  3.77 \\ \cline{2-11} 
 &
  RMD-INR &
  249 &
  \textbf{77.79} &
  -0.037 &
  0.384 &
  0.223 &
  0.421 &
  -5.95 &
  0.42 &
  0.28 \\ \cline{2-11} 
 &
  Det-INR &
  248 &
  72.29 &
  \textbf{-} &
  - &
  - &
  - &
  - &
  0.374 &
  \underline{\textbf{0.11}} \\ \hline
\multirow{4}{*}{\begin{tabular}[c]{@{}c@{}}Isabel\\ 500x500x100\\ 100MB\end{tabular}} &
  REV-INR &
  247 &
  \textbf{47.17} &
  \textbf{0.368} &
  \textbf{0.544} &
  \textbf{0.394} &
  \textbf{-1.668} &
  -1.01 &
  15.34 &
  \textbf{1.66} \\ \cline{2-11} 
 &
  MCD-INR &
  246 &
  45.09 &
  0.196 &
  0.238 &
  0.361 &
  40.45 &
  -3.539 &
  7.91 &
  88.55 \\ \cline{2-11} 
 &
  RMD-INR &
  249 &
  43.79 &
  0.065 &
  0.234 &
  0.319 &
  -0.195 &
  -3.462 &
  4.16 &
  2.92 \\ \cline{2-11} 
 &
  Det-INR &
  248 &
  \underline{\textbf{47.61}} &
  - &
  - &
  - &
  - &
  - &
  4.82 &
  \underline{\textbf{1.59}} \\ \hline
\multirow{4}{*}{\begin{tabular}[c]{@{}c@{}}Combustion\\ 480x720x120\\ 158.2MB\end{tabular}} &
  REV-INR &
  246 &
  \textbf{45.55} &
  \textbf{0.688} &
  \textbf{0.522} &
  \textbf{0.613} &
  -1.188 &
  -0.951 &
  10.87 &
  \underline{\textbf{2.75}} \\ \cline{2-11} 
 &
  MCD-INR &
  245 &
  42.68 &
  0.338 &
  0.303 &
  0.542 &
  \textbf{-5.172} &
  \textbf{-4.785} &
  5.23 &
  138.25 \\ \cline{2-11} 
 &
  RMD-INR &
  249 &
  43.01 &
  0.115 &
  0.25 &
  0.478 &
  0.47 &
  -4.51 &
  3.48 &
  4.84 \\ \cline{2-11} 
 &
  Det-INR &
  248 &
  \underline{\textbf{47.7}} &
  - &
  - &
  - &
  - &
  - &
  3.002 &
  \textbf{4.44} \\ \hline
\multirow{4}{*}{\begin{tabular}[c]{@{}c@{}}Foot\\ 500x500x360\\ 343.32MB\end{tabular}} &
  REV-INR &
  246 &
  \textbf{43.34} &
  \textbf{0.573} &
  \textbf{0.688} &
  \textbf{0.643} &
  \textbf{-2.337} &
  -1.551 &
  21.94 &
  \underline{\textbf{3.73}} \\ \cline{2-11} 
 &
  MCD-INR &
  245 &
  43.04 &
  0.266 &
  0.468 &
  0.614 &
  5.431 &
  \textbf{-3.617} &
  8.73 &
  997.02 \\ \cline{2-11} 
 &
  RMD-INR &
  249 &
  42.39 &
  0.134 &
  0.398 &
  0.545 &
  0.183 &
  -3.51 &
  8.13 &
  10.45 \\ \cline{2-11} 
 &
  Det-INR &
  248 &
  \underline{\textbf{44.83}} &
  - &
  - &
  - &
  - &
  - &
  9.93 &
  \textbf{5.85} \\ \hline
\multirow{4}{*}{\begin{tabular}[c]{@{}c@{}}Vortex\\ 512x512x512\\ 512MB\end{tabular}} &
  REV-INR &
  247 &
  \textbf{58.66} &
  \textbf{0.848} &
  \textbf{0.776} &
  0.348 &
  -2.869 &
  -1.979 &
  37.95 &
  \textbf{11.8} \\ \cline{2-11} 
 &
  MCD-INR &
  246 &
  53.05 &
  0.191 &
  0.113 &
  \textbf{0.46} &
  \textbf{-4.34} &
  -4.355 &
  15.93 &
  455.19 \\ \cline{2-11} 
 &
  RMD-INR &
  249 &
  58.39 &
  0.04 &
  0.305 &
  0.409 &
  -0.118 &
  \textbf{-4.89} &
  9.43 &
  15.52 \\ \cline{2-11} 
 &
  Det-INR &
  248 &
  \underline{\textbf{60.02}} &
  - &
  - &
  - &
  - &
  \textbf{-} &
  14.87 &
  \underline{\textbf{8.98}} \\ \hline
\multirow{4}{*}{\begin{tabular}[c]{@{}c@{}}Heptane\\ 512x512x512\\ 512MB\end{tabular}} &
  REV-INR &
  247 &
  \textbf{46.88} &
  \textbf{0.488} &
  \textbf{0.562} &
  \textbf{0.682} &
  -2.119 &
  -1.494 &
  21.14 &
  \underline{\textbf{5.56}} \\ \cline{2-11} 
 &
  MCD-INR &
  246 &
  43.84 &
  0.19 &
  0.446 &
  0.621 &
  \textbf{-5.6} &
  \textbf{-5.371} &
  15.96 &
  455.74 \\ \cline{2-11} 
 &
  RMD-INR &
  249 &
  41.22 &
  0.145 &
  0.503 &
  0.663 &
  0.419 &
  -5.06 &
  8.84 &
  15.58 \\ \cline{2-11} 
 &
  Det-INR &
  248 &
  \underline{\textbf{47.35}} &
  - &
  - &
  - &
  - &
  - &
  14.59 &
  \textbf{8.84} \\ \hline
\end{tabular}%
}
\end{table*}

\begin{figure}[!tb]
\centering
\begin{subfigure}[t]{0.48\textwidth}
    \centering
    \includegraphics[width=\textwidth]{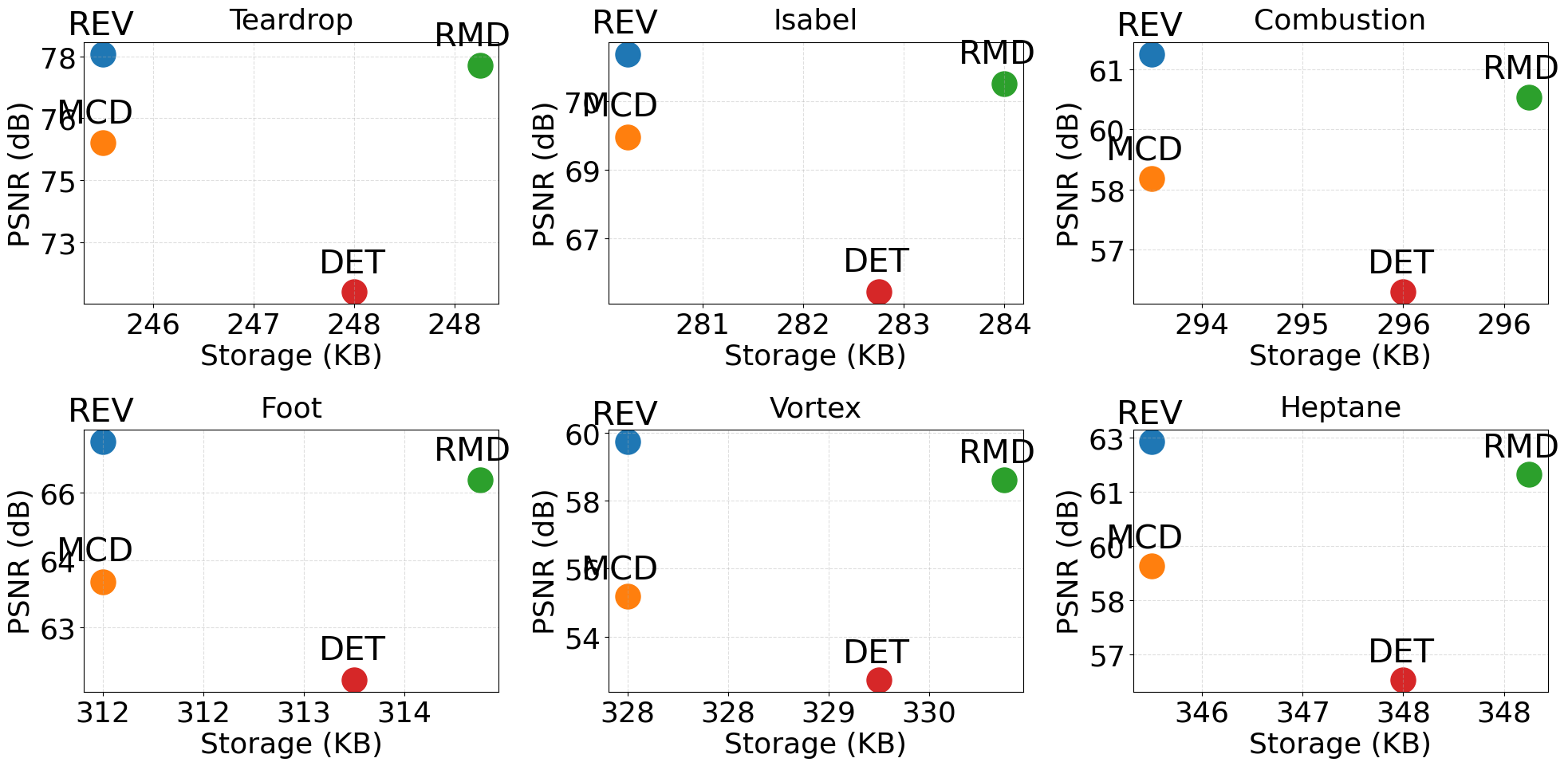}
    \caption{PSNR vs. Storage trade-off scatterplot.}
    \label{storage_vs_psnr}
\end{subfigure}
\vskip 8pt
\begin{subfigure}[t]{0.48\textwidth}
    \centering
    \includegraphics[width=\textwidth]{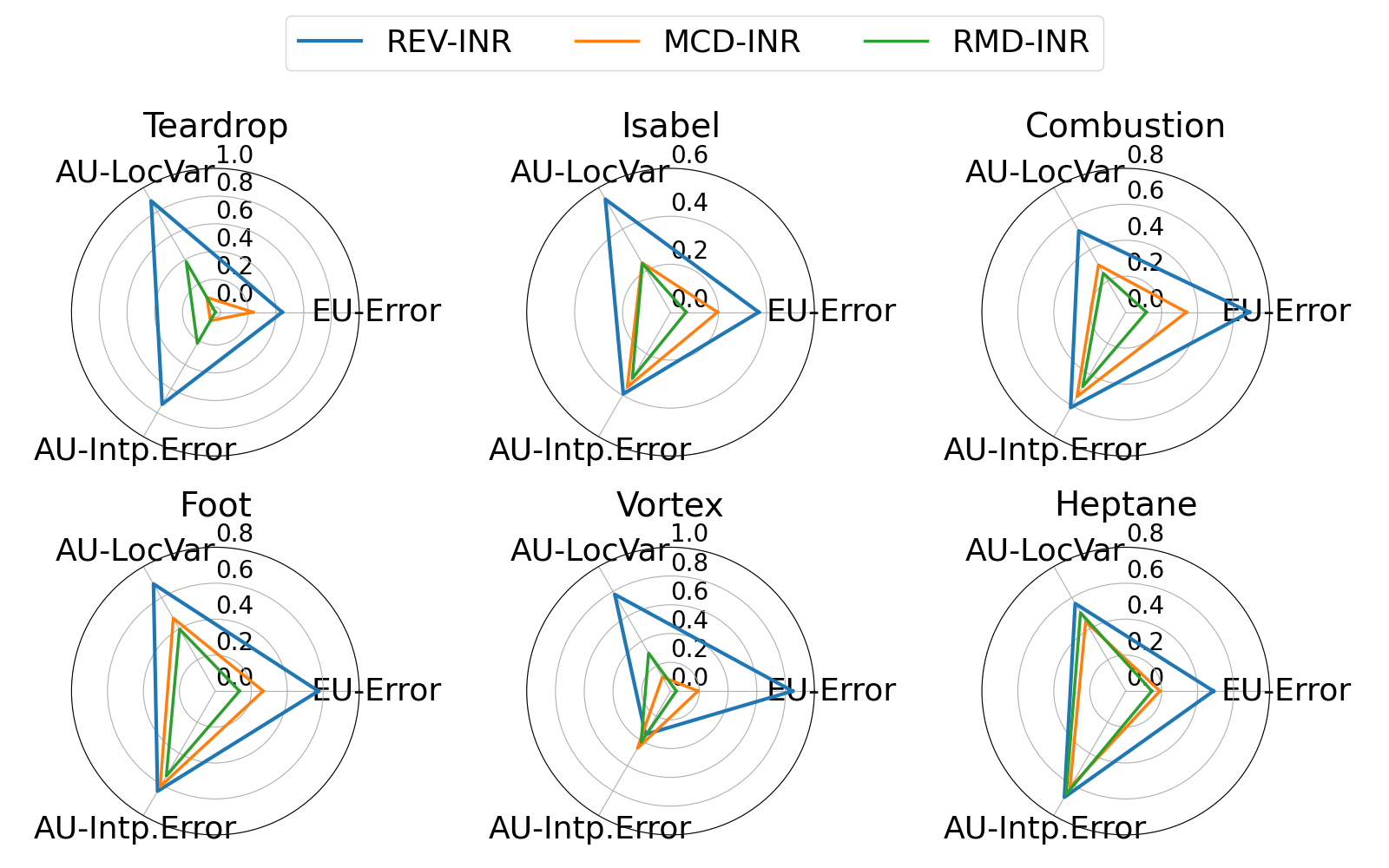}
    \caption{Correlation strength Rader plot.}
    \label{rader_plot}
\end{subfigure}
\caption{%
\rmark{Fig.~\ref{storage_vs_psnr} shows that REV-INR has the best trade-off as it is located at the top-left corner indicating minimal storage and higher PSNR among all methods. Fig.~\ref{rader_plot} depicts that REV-INR demonstrates the highest correlation between AU-LocalVar., AU-Interp. Error, and EU-prediction error for all methods.}}
\label{metric_plot}
\end{figure}

\textbf{Model Size vs. Reconstruction.}
Table~\ref{inr_results} presents the quantitative evaluation of the volume reconstruction quality for comparable model sizes. To assess the overall volume reconstruction quality, we compute the peak signal-to-noise ratio (PSNR). We observe that while REV-INR produces the highest PSNR among the uncertain INRs,  for the Combustion, Isabel, Foot, Vortex, and Heptane datasets, REV-INR produces either comparable or slightly lower PSNR compared to Det-INR. For the Teardrop dataset, REV-INR produces the highest PSNR. It is to be noted that while Det-INR produces slightly higher PSNR for a few  test data, in Section~\ref{uncert_isosurf}, we have demonstrated how Det-INR can produce erroneous isosurfaces even with higher PSNR. Hence, we advocate for uncertainty-aware INRs over their deterministic counterparts to enhance reliability in INR-predicted results. Among the  uncertain INRs, REV-INR produces superior results, achieving up to a compression ratio of 2100$\times$. \rmark{This superior trade-off between storage and PSNR is also evident in Fig.~\ref{storage_vs_psnr}, where REV-INR consistently occupies the top-left region of the scatter plot, indicating a better storage–PSNR trade-off.}

\textbf{Uncertainty Estimation.}
To assess the interpretability and usability of AU and EU values, we compute the Pearson correlation between the EU and error fields to evaluate their alignment. To demonstrate the quality of the AU values, we compute the  correlation of AU with (1) local variance and (2) interpolation-based  error, obtained by first down-sampling a volume and then up-sampling it back to the original resolution using linear interpolation. We observe that,  AU estimates generated by REV-INR yield the highest correlation with local variance values. When computing correlations with interpolation error, REV-INR achieves the highest correlation for all datasets except the Vortex dataset, where MCD-INR produces the highest correlation. This higher correlation between AU-LocalVar., AU-Interp. Error, and EU-prediction error is also evident in Fig.~\ref{rader_plot}, where the Radar plot demonstrates the same. Another way to evaluate uncertainty is by computing the Negative Log-Likelihood (NLL) using the predicted mean and variance of AU and EU to parameterize Gaussian distributions as suggested in~\cite{Tianyu}. Table~\ref{inr_results} reports the results. REV-INR achieves the best NLL EU on the Isabel and Foot datasets, while MCD-INR performs best on the remaining datasets. For NLL AU, RMD-INR and MCD-INR yields lower NLL than REV-INR, which is expected since both methods are explicitly optimized using a Gaussian NLL loss.

\textbf{Training and Inference Timings.}
From Table~\ref{inr_results}, we observe that REV-INR takes the longest to train, with training time reaching up to 2–3$\times$ that of the other uncertain INRs, but its inference time either the fastest or is on par with Det-INR. For the Combustion, Foot, and Heptane datasets, REV-INR achieves the fastest inference. The primary reason for REV-INR's longer training time is that it uses a complex evidential loss function, which computes the KL divergence between two NIG distributions—a computationally expensive operation. Additionally, the AU and EU regularization introduces further computations, contributing to the higher training cost of REV-INR. However, now that training is performed offline and is a one-time operation while inference time is more critical for visual analysis. As REV-INR is either the fastest or comparable to Det-INR in inference time with robust uncertainty estimates, we believe that REV-INR can be considered the preferred choice for uncertainty-aware volume modeling and visualization.

\textbf{Ablation Study: REV-INR With and Without Uncertainty Regularization.}
We conduct an ablation study to evaluate the effectiveness of the proposed EU and AU regularization losses. We train REV-INR with and without the regularization on Teardrop, Isabel, Foot, and Heptane datasets. The results are reported in Table~\ref{ablation_study}. We observe that, without the uncertainty regularization, REV-INR yields lower correlations, making the interpretation of the unregularized uncertainty estimates more challenging. Hence, we conclude that the proposed uncertainty regularizations are essential for training REV-INR to obtain meaningful uncertainty estimates.

\begin{table}[!t]
\caption{Impact of EU and AU regularization for REV-INR.}
\label{ablation_study}
\resizebox{\columnwidth}{!}{%
\begin{tabular}{|c|cc|cc|cc|}
\hline
\multirow{2}{*}{\textbf{Dataset}} &
  \multicolumn{2}{c|}{\textbf{Corr. (EU and Prediction Error) $\uparrow$}} &
  \multicolumn{2}{c|}{\textbf{Corr. (AU and  Loc. Var.) $\uparrow$}} &
  \multicolumn{2}{c|}{\textbf{Corr. (AU and Interp. Err.) $\uparrow$}} \\ \cline{2-7} 
 &
  \multicolumn{1}{c|}{\textbf{\begin{tabular}[c]{@{}c@{}}With \\ Regularization\end{tabular}}} &
  \textbf{\begin{tabular}[c]{@{}c@{}}Without\\ Regularization\end{tabular}} &
  \multicolumn{1}{c|}{\textbf{\begin{tabular}[c]{@{}c@{}}With \\ Regularization\end{tabular}}} &
  \textbf{\begin{tabular}[c]{@{}c@{}}Without\\ Regularization\end{tabular}} &
  \multicolumn{1}{c|}{\textbf{\begin{tabular}[c]{@{}c@{}}With \\ Regularization\end{tabular}}} &
  \textbf{\begin{tabular}[c]{@{}c@{}}Without\\ Regularization\end{tabular}} \\ \hline
Teardrop &
  \multicolumn{1}{c|}{0.445} &
  0.317 &
  \multicolumn{1}{c|}{0.889} &
  0.351 &
  \multicolumn{1}{c|}{0.73} &
  0.192 \\ \hline
Foot &
  \multicolumn{1}{c|}{0.573} &
  0.04 &
  \multicolumn{1}{c|}{0.688} &
  0.119 &
  \multicolumn{1}{c|}{0.643} &
  0.291 \\ \hline
Isabel &
  \multicolumn{1}{c|}{0.368} &
  0.012 &
  \multicolumn{1}{c|}{0.544} &
  0.077 &
  \multicolumn{1}{c|}{0.394} &
  0.122 \\ \hline
Heptane &
  \multicolumn{1}{c|}{0.488} &
  0.07 &
  \multicolumn{1}{c|}{0.562} &
  0.163 &
  \multicolumn{1}{c|}{0.682} &
  0.391 \\ \hline
\end{tabular}
}
\end{table}

\section{Discussion and Limitations}

Our results establish the overall superiority of REV-INR, producing calibrated and interpretable uncertainty fields, allowing robust value and uncertainty prediction within a single forward pass. We also observe that while Deterministic INRs (Det-INR) occasionally report slightly higher PSNR, they can produce erroneous or topologically incorrect isosurfaces, resulting in misleading interpretations. This observation emphasizes that PSNR is only a global quality metric and does not reflect structural correctness of extracted isosurfaces. In contrast, uncertainty-aware models such as REV-INR produce interpretable, topologically correct results, underscoring the necessity of uncertainty modeling for reliable INR-based visualization.

A potential limitation of REV-INR is its higher training cost, primarily due to the evidential loss and uncertainty regularization computations. However, since training is a one-time offline process, and in practice, faster volumes reconstruction is more desirable, and considering REV-INR's overall superiority over existing techniques, REV-INR can be the preferred choice for uncertainty-aware volume modeling. Our future research will explore multi-resolution hash encoding~\cite{hash_encoding1, hash_encoding2}, feature-grid-based INR construction, and approximate KL-divergence-based loss computations to reduce the training time of REV-INR while retaining the advantages of REV-INR.

While we find that using the gradient as a regularization measure for aligning the AU estimates works well for all the test volume datasets, there may be volumetric datasets where the gradient is only weakly correlated with data variance. In such cases, the estimated AU may not generalize well to capture the local data variance. Finally, across all three methods, uncertainty regularization introduces a trade-off between reconstruction fidelity and uncertainty calibration. Future work will explore automated strategies to balance these competing objectives, making REV-INR more robust while improving reconstruction quality without sacrificing reliability.

\section{Conclusions and Future Work}

This work highlights the importance of incorporating uncertainty into INR-based visual analysis. We introduce REV-INR, an uncertainty-aware regularized INR, and demonstrate its superiority over two uncertain INRs. Future work will focus on accelerating REV-INR and extending it to multivariate and ensemble datasets. These uncertainty-informed visualizations reveal regions requiring further training and expose model limitations, underscoring the importance of model confidence for trust and reliability in scientific decision-making.

\section*{Acknowledgments}
This work was supported by IIT Kanpur's Initiation Grant (project number $IITK /CS /2022307$). We thank the anonymous reviewers for providing insightful comments that helped in improving the paper. The Hurricane Isabel dataset has kindly been provided by Wei Wang, Cindy Bruyere, Bill Kuo, and others at NCAR. Tim Scheitlin at NCAR converted the data into the Brick-of-Float format described above. The Combustion dataset is made available by Dr. Jacqueline Chen at Sandia Laboratories through US Department of Energy’s SciDAC Institute for Ultrascale Visualization. The Foot dataset is courtesy of Philips Research, Hamburg, Germany and was obtained from The Volume Library (http://volume.open-terrain.org/). We thank the University of Utah Center for Simulation of Accidental Fires and Explosions for making the Heptane dataset available. 

%This manuscript has been authored by UT Battelle, LLC under Contract No. DE-AC05-00OR22725 with the U.S. Department of Energy. The publisher, by accepting the article for publication, acknowledges that the U.S. Government retains a non-exclusive, paid up, irrevocable, world-wide license to publish or reproduce the published form of the manuscript, or allow others to do so, for U.S. Government purposes. The DOE will provide public access to these results in accordance with the DOE Public Access Plan (http://energy.gov/downloads/doe-public-access-plan).

\bibliographystyle{abbrv-doi-hyperref}

\bibliography{template}

\end{document}